\documentclass[acmtog,nonacm]{acmart}
\acmSubmissionID{710}
\usepackage{pifont}
\usepackage{float}
\usepackage{subcaption}
\AtBeginDocument{%
  }

\setcopyright{acmlicensed}
\copyrightyear{2026}
\acmYear{2026}
\acmDOI{XXXXXXX.XXXXXXX}

\acmVolume{1}
\acmNumber{1}
\acmArticle{1}
\acmMonth{1}
\begin{document}
\title{LOME: Learning Human-Object Manipulation with Action-Conditioned Egocentric World Model \\ }

\author{Quankai Gao}
% \authornote{Both authors contributed equally to this research.}
\email{quankaig@usc.edu}
% \orcid{1234-5678-9012}
% \author{G.K.M. Tobin}
% \authornotemark[1]
% \email{webmaster@marysville-ohio.com}
\affiliation{%
  \institution{University of Southern California}
  \city{Los Angeles}
  % \state{Ohio}
  \country{USA}
}

\author{Jiawei Yang}
% \authornote{Both authors contributed equally to this research.}
\email{yangjiaw@usc.edu}
% \orcid{1234-5678-9012}
% \author{G.K.M. Tobin}
% \authornotemark[2]
% \email{webmaster@marysville-ohio.com}
\affiliation{%
  \institution{University of Southern California}
  \city{Los Angeles}
  % \state{Ohio}
  \country{USA}
}

\author{Le Chen}
% \authornote{Both authors contributed equally to this research.}
\email{le.chen@tuebingen.mpg.de}
% \orcid{1234-5678-9012}
% \author{G.K.M. Tobin}
% \authornotemark[3]
% \email{webmaster@marysville-ohio.com}
\affiliation{%
  \institution{Max Planck Institute for Intelligent Syetems}
  \city{Tübingen}
  % \state{Ohio}
  \country{Germany}
}

\author{Qiangeng Xu}
% \authornote{Both authors contributed equally to this research.}
\email{charlie.learning@yahoo.com}
% \orcid{1234-5678-9012}
% \author{G.K.M. Tobin}
% \authornotemark[3]
% \email{webmaster@marysville-ohio.com}
\affiliation{%
  \institution{Waymo}
  \city{Mountain View}
  % \state{Ohio}
  \country{USA}
}

\author{Yue Wang}
% \authornote{Both authors contributed equally to this research.}
\email{yue.w@usc.edu}
% \orcid{1234-5678-9012}
% \author{G.K.M. Tobin}
% \authornotemark[3]
% \email{webmaster@marysville-ohio.com}
\affiliation{%
  \institution{University of Southern California}
  \city{Los Angeles}
  % \state{Ohio}
  \country{USA}
}

\renewcommand{\shortauthors}{Trovato et al.}

\renewcommand{\shortauthors}{Gao et al.}

%%
%% The abstract is a short summary of the work to be presented in the
%% article.
\begin{abstract}
Learning human-object manipulation presents significant challenges due to its fine-grained and contact-rich nature of the motions involved. Traditional physics-based animation requires extensive modeling and manual setup, and more importantly, it neither generalizes well across diverse object morphologies nor scales effectively to real-world environment. To address these limitations, we introduce LOME, an egocentric world model that can generate realistic human-object interactions as videos conditioned on an input image, a text prompt, and per-frame human actions, including both body poses and hand gestures. LOME injects strong and precise action guidance into object manipulation by jointly estimating spatial human actions and the environment contexts during training. After finetuning a pretrained video generative model on videos of diverse egocentric human-object interactions, LOME demonstrates not only high action-following accuracy and strong generalization to unseen scenarios, but also realistic physical consequences of hand–object interactions, e.g., liquid flowing from a bottle into a mug after executing a ``pouring'' action. Extensive experiments demonstrate that our video-based framework significantly outperforms state-of-the-art image-based and video-based action-conditioned methods and Image/Text-to-Video (I/T2V) generative model in terms of both temporal consistency and motion control. LOME paves the way for photorealistic AR/VR experiences and scalable robotic training, without being limited to simulated environments or relying on explicit 3D/4D modeling. Our
project page is available at: https://zerg-overmind.github.io/LOME.github.io/.
\end{abstract}

%%
%% The code below is generated by the tool at http://dl.acm.org/ccs.cfm.
%% Please copy and paste the code instead of the example below.
%%
% \ccsdesc[500]{Computing methodologies~Video}

%%
%% Keywords. The author(s) should pick words that accurately describe
%% the work being presented. Separate the keywords with commas.

% \keywords{Video}

% \received{20 February 2007}
% \received[revised]{12 March 2009}
% \received[accepted]{5 June 2009}

%%
%% This command processes the author and affiliation and title
%% information and builds the first part of the formatted document.
%\input{derekPre}
\begin{teaserfigure}
    \centering
    \includegraphics[width=\linewidth]{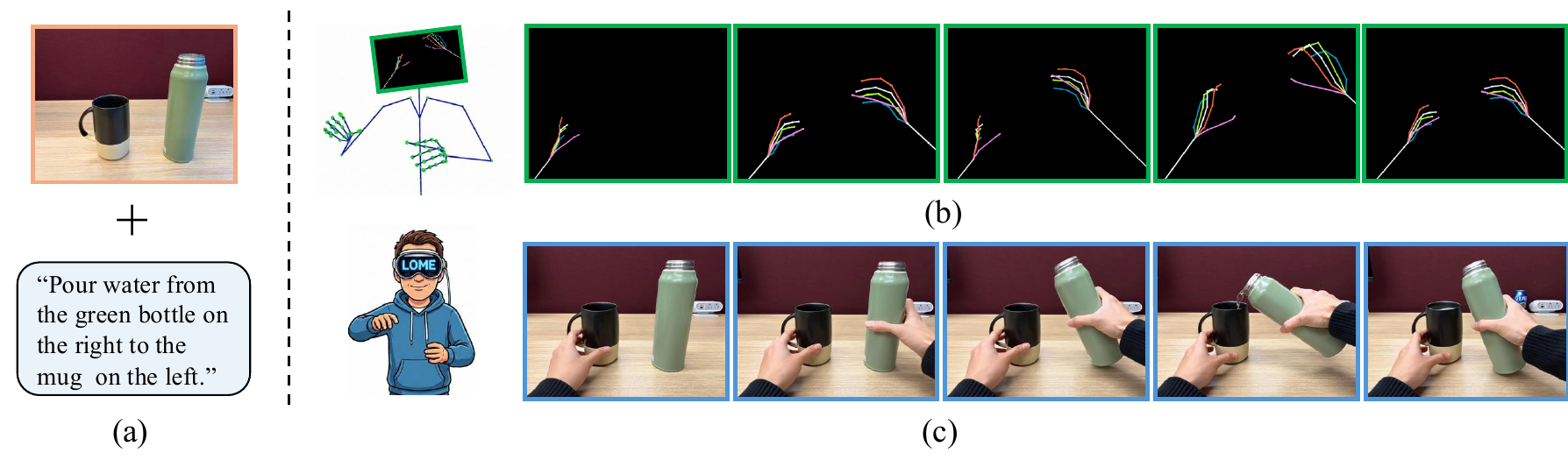}
    \vspace{-5mm}
    \caption{\textbf{Overview}. Given a reference image and a text instruction describing the manipulations shown in (a), \textbf{LOME} generates a temporally consistent egocentric hand–object interaction video, as shown in (c), conditioned on the corresponding per-frame human actions as in (b). 
    % We demonstrate generative results in in-the-wild environments with real objects captured in our lab. 
    Beyond accurate action adherence, LOME synthesizes realistic physical consequences of hand–object interactions, such as liquid dynamics when pouring from a bottle into a mug.}
    \label{fig:teaser}
\end{teaserfigure}

\maketitle

\section{Introduction}

% Learning human–object manipulation is critical for computer vision, graphics, and robotics. It requires modeling physical dynamics, contact-rich motion, and the causal relationship between human actions and object motions, $i.e.$ how object motions evolve in response to human actions. Therefore, human-object manipulation entails more than hand-object interaction (HOI) detection~\cite{zhang2023exploring,xie2022chore,park2023viplo}, 3D/4D hand–object reconstruction~\cite{zhang2020perceiving,xie2024rhobin,wen2025efficient,ye2023diffusion}, or human action prediction conditioned on static or moving objects~\cite{ye2023affordance,corona2020ganhand,cao2021reconstructing,li2023object}. These reconstruction-based methods are inherently limited in their ability to synthesize novel interactions and often fail to generalize to unseen environments. 
% Recent works~\cite{wu2025human, xu2024interdreamer} have explored long-horizon HOI synthesis using large language models, but these approaches remain largely constrained to simulated environments. Consequently, there is a critical need for a general and scalable pipeline that supports physically plausible manipulation across diverse real-world settings, while eliminating the need for non-scalable per-scene optimization.

Learning human–object manipulation is critical for computer vision, graphics, and robotics, as it requires modeling physical dynamics, contact-rich motion, and the causal relationship between human actions and object motions, $i.e.$ how object motions evolve in response to human actions. Therefore, human-object manipulation entails more than hand-object interaction (HOI) detection~\cite{zhang2023exploring,xie2022chore,park2023viplo}, 3D/4D hand–object reconstruction from a single image or a video clip~\cite{zhang2020perceiving,xie2024rhobin,wen2025efficient,ye2023diffusion}, or human action prediction given either static~\cite{ye2023affordance,corona2020ganhand,cao2021reconstructing} or moving objects~\cite{li2023object} as reconstruction-based methods are inherently limited in their ability to synthesize novel interactions, often failing to generalize to unseen environments. 
Recent works~\cite{wu2025human, xu2024interdreamer} have demonstrated progress in synthesizing HOI with long-context reasoning and planning by large language models. However, these approaches remain largely constrained to simulated environments. Consequently, there is a critical need for a general pipeline capable of handling diverse objects and environments. Such a system must simulate physically plausible interactions for realistic object manipulation, while eliminating the need for non-scalable per-scene optimization.

Video diffusion models offer a promising alternative. Pretrained on large-scale data, they capture rich motion priors and generalize across diverse dynamics~\cite{Veo, Runaway, Luma, wan2025wan, yang2024cogvideox}. This makes them natural candidates for world models~\cite{ha2018world,bruce2024genie}. However, conditioning on text or images alone produces poor motion quality—a limitation that data scaling does not resolve~\cite{kang2024far,chefer2025videojam}. Adding explicit spatial control improves motion synthesis~\cite{chefer2025videojam,burgert2025go,geng2025motion,shin2025motionstream}, but existing control signals like optical flow or motion trajectories are problematic for manipulation: they suffer from occlusions and, more fundamentally, they prescribe object motion rather than letting it emerge from human actions.

Motivated by these insights, we propose incorporating human actions as explicit spatial control into a pretrained video model to learn general hand–object interactions. Leveraging rich motion priors acquired during large-scale pretraining, the video model can learn spatial correspondences between action conditions and human-object interactions with lightweight fine-tuning, avoiding heavy task-specific retraining.
% To further reduce view–action variability, we constrain our framework to an egocentric setting, which more closely reflects how humans perform daily tasks. 
To more closely reflect how humans perform daily tasks, we initiate our framework in an egocentric setting. Instead of learning a discrete and fix-sized latent action space during training and generating interactive videos through inference-time action retrieval as recent video-based world models~\cite{he2025matrix,che2024gamegen}, our method handles continuous human actions by conditioning on: 1) an input image that specifies the environment and the objects; 2) a text prompt that briefly describes the intended human-object interactions; 3) per-timestep human actions including body poses and hand gestures.  

Moreover, our empirical results show that naïvely using actions as conditioning signals when training diffusion models often fails to produce precise, realistic hand–object interactions in generated videos.
Drawing inspiration from VideoJAM~\cite{chefer2025videojam}, we reformulate the training objective of the video diffusion model to explicitly model the joint action-environment distribution by denoising the concatenation of latent representations of action maps and generated videos. 
Learning this joint distribution facilitates a cleaner decoupling between actions and environment dynamics, enabling the model to generalize more effectively to novel actions and unseen environments. 

To demonstrate the effectiveness of LOME, we fine-tune a pretrained video diffusion model (Wan2.1-VACE-14B~\cite{wan2025wan}) with lightweight LoRA adaptation. On EgoDex dataset and in-the-wild samples, LOME achieves 66.85\% PCK@20 for action-following accuracy versus 51.33\% for the best baseline, and improves FVD from 59.83 to 39.58. In our user studies, LOME receives 97\% preference for action-following and 94\% for visual quality. Beyond metrics, LOME synthesizes realistic physical consequences without 3D/4D reconstruction and simulation.

Our contributions can be summarized as follows:
\begin{itemize}
    \item We introduce \textbf{LOME}, an action-conditioned egocentric world model that learns fine-grained, contact-rich human-object interactions from real-world video captures.
    \item We propose learning the joint distribution of both actions and environmental contexts (including objects) with multiple conditions, achieving precise and realistic manipulation.
    \item We demonstrate that our method can simulate coherent human-object interactions across diverse real-world scenarios, including interactions with multiple objects, with realistic physical consequences.
    
    % Our generative outputs reflect causal dynamics during interactions. For instance, liquid naturally falls out of a cup following a “pouring” action.
\end{itemize}

\begin{figure*}[h]
    \centering
    \includegraphics[width=\linewidth]{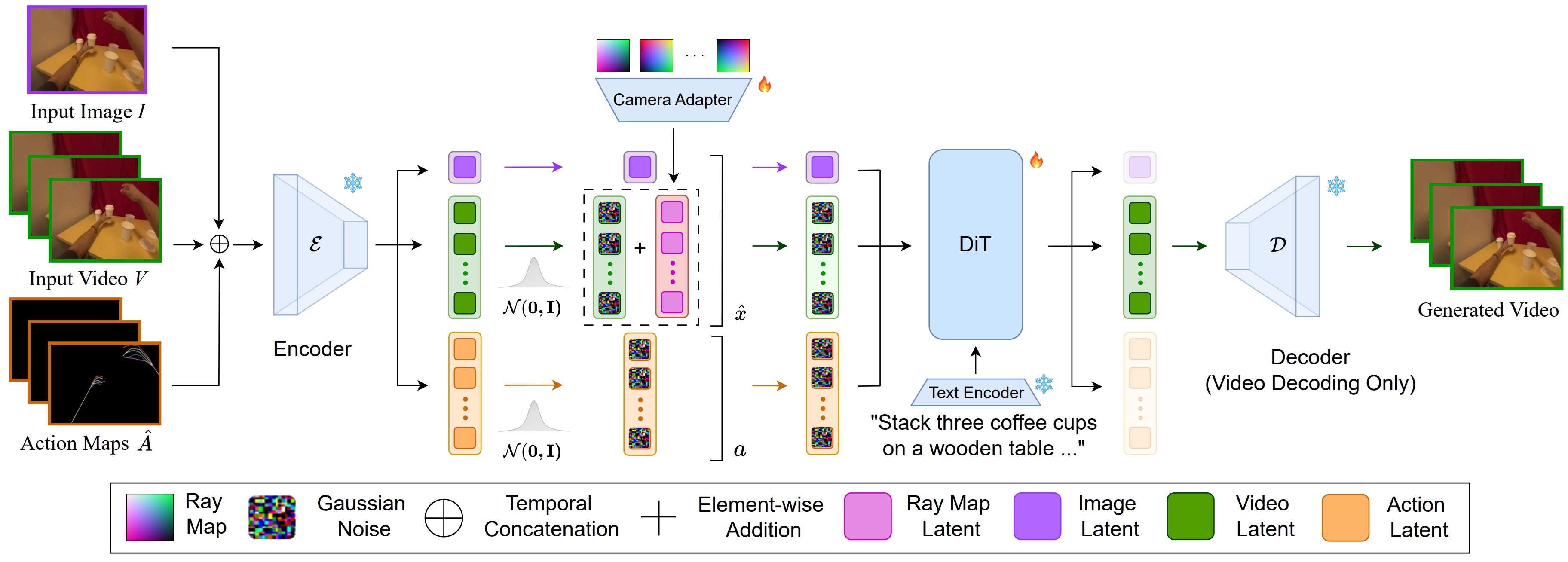}
    \vspace{-5mm}
    \caption{\textbf{Training pipeline of LOME.} A pretrained VAE encoder $\mathcal{E}$ maps the reference image $I$, input video $V$, and rasterized 2D action maps $\hat{A}$ to latent representations. A camera adapter encodes per-frame ray maps into camera features, which are added to the video latents. A Diffusion Transformer (DiT), conditioned on a text prompt, denoises the concatenated noisy action and video latents, and a pretrained decoder $\mathcal{D}$ reconstructs the generated video.}
    \label{fig:pipeline}
    % \vspace{-2mm}
\end{figure*}
\section{Related Work}
\subsection{Human–object manipulation}
Existing research on human–object manipulation or hand–object interactions primarily relies on human and object detection~\cite{bambach2015lending,mittal2011hand,shan2020understanding,kwon2021h2o}, pose estimation, and fitting~\cite{pavlakos2019expressive,boukhayma20193d,kulon2020weakly,li2023ego, zhang2025bimart} from images, with 3D parametric models or object templates~\cite{romero2022embodied, loper2023smpl,sun2018pix3d,chen2025sam}.  Beyond these reconstruction-based approaches, generative methods have emerged to predict visual affordances~\cite{ye2023affordance,corona2020ganhand,karunratanakul2020grasping,lu2024ugg, prakash2025synthesizing, zhang2024hoidiffusion} or the future state given action signal~\cite{sudhakar2024controlling}. Recently, multi-modal models have facilitated fine-grained manipulation~\cite{wei2024grasp,zhong2025dexgrasp,li2024controllable,lai2024lego,christen2024diffh2o}, a capability essential for robotics. Moreover, learning object manipulation is especially crucial in robotics, enabling humanoid robots to imitate human behavior while interacting with objects accurately and safely. For instance, Vision-Language-Action (VLA) models~\cite{kim2024openvla,amin2025pi,zhou2025vision} allow robotic agents to directly regress action sequences by training on action-visual-language triplets. While approaches like DexWM~\cite{goswami2025world} learn manipulation from egocentric videos, they tend to focus on predicting future states rather than following explicit action conditions.

In this work, we learn human–object manipulation using a video generative model conditioned on human actions, mirroring how humans explore and interact with the world, but without explicit 3D/4D reconstruction. Action-conditioned video generation has been witnessed in navigation~\cite{bai2025whole} and hand/agent-object interaction~\cite{wang2025precise, xie2026generated, tu2025playerone, xue2024hoi, fan2025re}.

\subsection{Video Generative Model}
Most video generative models are pretrained with multiple different conditions, such as text  (T2V), image (I2V), and action~\cite{bai2025whole,he2025matrix, bruce2024genie,li2025hunyuan,tu2025playerone}, and more~\cite{hu2024animate,chang2023magicpose,jiang2025vace,he2024cameractrl,wang2024motionctrl}, to enable controllable video generation via diffusion process.
Early works~\cite{blattmann2023align,ho2022imagen,hong2022cogvideo,singer2022make} focused on extending image diffusion models with additional temporal modules to produce videos. More recently, the diffusion backbone of many video generative models has converged toward the Diffusion Transformer (DiT) architecture~\cite{peebles2023scalable}. Beyond architectural advances, scaling training data has become increasingly critical~\cite{blattmann2023stable} for improving video generation quality, a trend clearly demonstrated by recent closed-~\cite{OpenAI,Veo,Runaway,Luma} and open-source models~\cite{wan2025wan,kong2024hunyuanvideo,yang2024cogvideox, hong2022cogvideo} and consistent with ``The Bitter Lesson"~\cite{Sutton2025the}. By training on a vast amount of data, video generative models show impressive abilities of dynamic synthesis and spatial understanding~\cite{bar2025navigation} and abstract reasoning~\cite{wiedemer2025video}. 

Despite these advances, motion generation and control in video models remain challenging. Recent findings~\cite{chefer2025videojam,kang2024far} suggest that data scaling alone is insufficient to capture realistic motion dynamics or physical interactions. Furthermore, post-training strategies like Reinforcement Learning from Human Feedback (RLHF)~\cite{wu2025densedpo,xue2025dancegrpo,cai2025phygdpo} do not necessarily guarantee better performance ($e.g.$ motion synthesis). This is due to the limited capabilities of base models~\cite{yue2025does} and ambiguous human-preference rewards, which can often induce undesirable behaviors such as slow motion or other forms of motion degradation~\cite{wu2025densedpo}. Instead of improving the base models via data scaling or RLHF, model finetuning with spatial visual prompts for controllable generation often leads to more effective and controllable motion synthesis~\cite{burgert2025go,geng2025motion,shin2025motionstream}.

Based on these insights, we integrate human actions as explicit spatial control signals into the video generative framework using action-conditioned fine-tuning.

% \subsection{Multi-condition Joint Optimization}

\section{LOME}
Our goal is to generate fine-grained, contact-rich human-object manipulation videos. Given an input image, a text description, and per-frame human actions, we want the generated video to: (1) follow the specified actions precisely, (2) depict physically plausible object responses, and (3) generalize to novel actions and unseen environments. We build on pretrained video diffusion models and introduce two key designs: \textit{spatial action maps} for precise control and \textit{joint action-environment modeling} for enforcing consistency. In this section, we first review preliminaries, then detail our approach.

\subsection{Preliminary}
Given a condition $c$, a condition-to-video (c2V) diffusion model 
generates the corresponding video $V\in\mathbb{R}^{3\times L\times H\times W}$, where $L$ is the number of temporal frames and $W\times H$ is the spatial resolution of each frame. 
Similar to image latent diffusion models~\cite{rombach2022high}, video diffusion models typically operate in a latent space defined by a
pretrained VAE with encoder $\mathcal{E}$ and decoder $\mathcal{D}$.
Specifically, a video $V$ is encoded as
$x = \mathcal{E}(V) \in \mathbb{R}^{C \times L' \times H' \times W'}$, where
$L', H', W'$ are temporally and spatially downsampled dimensions, and can be reconstructed by $V = \mathcal{D}(x)$. 
% Specifically, the entire video $V$ is divided into multiple chunks $V=\{V_i \mid i=0, ..., \frac{L-1}{d}\}$ along temporal dimension, where each chunk contains $d$ frames. The model then employs a diffusion module parameterized by $\theta$ to predict velocity $v_{\theta,t}(x \mid x_1, c)$ using flow matching~\cite{lipman2022flow} with the following objective:
The diffusion model, parameterized by $\theta$, learns to predict the velocity field $v_{\theta,t}(x \mid x_1, c)$ using flow matching~\cite{lipman2022flow}. The training objective is
\begin{equation}
\mathcal{L} =
\mathbb{E}_{t, q(x_0), p_t(x\mid x_0,c)}
\left[
\left\| v_{\theta,t}(x \mid x_0, c) - v_t(x \mid x_1, x_0, c) \right\|_2^2
\right],
\end{equation}
given a denoising timestep $ t \sim \mathcal{U}(0,1)$, a clean video latent $x_0 \sim q(x_0)$, a noisy video latent $x\sim p_t(x\mid x_0, c)$ and Gaussian noise $x_1\sim \mathcal{N}(0, \mathbf{I})$. The target velocity $v_t(x \mid c)$ is:
\begin{equation}
v_t(x \mid x_1, x_0, c) = \frac{dx_t}{dt} = x_1 - x_0,  \quad    x_t = (1-t) x_0 + tx_1.
\end{equation}
By omitting $x_1$ and $x_0$ for simplicity, Classifier-Free Guidance (CFG)~\cite{ho2022classifier} in inference is expressed as:
\begin{equation}
\label{original_cfg}
 v_{\theta,t}(x) = (1+w)\cdot v_{\theta,t}(x \mid c) - w\cdot v_{\theta,t}(x \mid \varnothing)
\end{equation}

\begin{figure}[t]
    \centering
    % First image
    \begin{minipage}[t]{1.0\linewidth} % Adjust width as needed
        \centering
        \includegraphics[width=\linewidth]{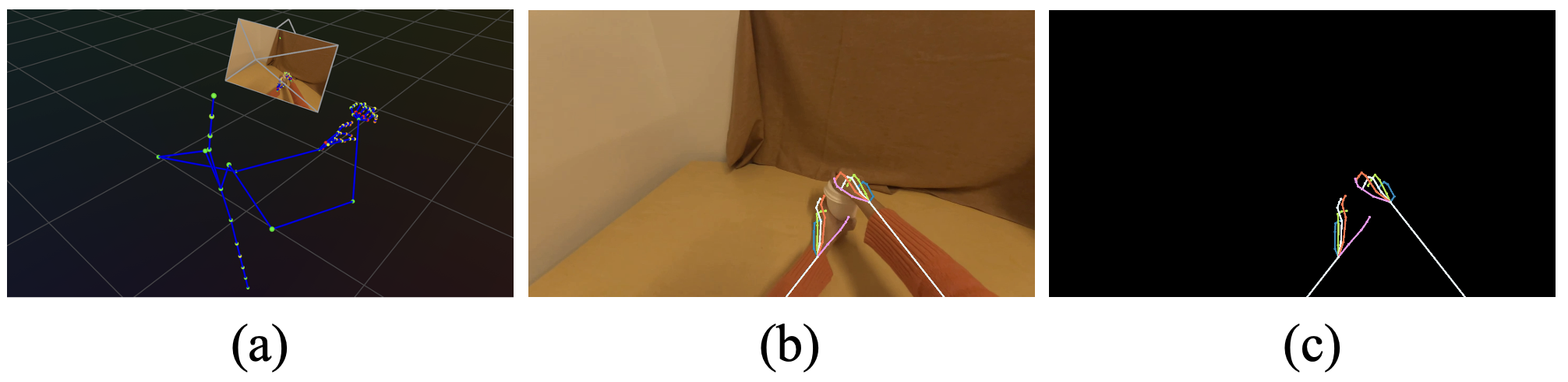} 
        % fig/tsne_2_crop (1).pdf
    \end{minipage}

    \vspace{-3mm}
    \caption{
       \textbf{Action conditioning at frame $i$.} 
       % The 3D human pose during video capture is shown in \textbf{(a)}; the projected human pose $A_i$ is visualized in \textbf{(b)}, obtained by filtering out keypoints and skeletons outside the camera frustum; and the resulting background-free 2D action map $\hat{A}_i$ is shown in \textbf{(c)}.
       (a) 3D human pose during video capture. (b) Projected 2D human pose $A_i$ after filtering out keypoints and skeleton segments outside the camera frustum. (c) Background-masked rasterized 2D action map $\hat{A}_i$ used as the conditioning signal.
    }
   \vspace{-2mm}
    \label{fig:action_rep}
\end{figure}

\subsection{Model Design}

We show the overview of LOME in Fig.~\ref{fig:pipeline} and describe details below. \\
\textbf{Inputs and Conditions.} 
LOME is conditioned on: text tokens $y\in\mathbb{R}^{d_1}$ describing the manipulation, an input image $I\in\mathbb{R}^{3\times H\times W}$ defining the scene, per-timestep camera extrinsics $\zeta\in\mathbb{R}^7$, and the action sequence $A\in\mathbb{R}^{d_2}$. 
To preserve scene content, we concatenate the encoded input image $x^{I} = \mathcal{E}(I)$ with the video latent $x$ along the temporal dimension to construct the \textit{environment latent} $[x^{I}, x]\in\mathbb{R}^{C\times (L'+1)\times H'\times W'}$, where $[\cdot,\cdot]$ denotes concatenation. The image latent $x^{I}$ remains noise-free during training, serving as a clean anchor for the scene.
To condition on camera poses, we encode camera extrinsics $\zeta$ and intrinsics $K$ into per-frame ray maps (Pl\"ucker embeddings) and pass them through a lightweight adapter $\mathcal{C}(\cdot)$ to obtain camera features $z=\mathcal{C}(\zeta, K)$, following~\cite{he2025cameractrl}. These are added element-wise to the video latent tokens $x$ to obtain the \textit{camera-conditioned environment latent} $\hat{x} =[x^{I}, x + z]$. \\
\textbf{Action Maps.} A key design question is how to represent actions. Recent work shows that explicit spatial control signals are more effective than latent representations for fine-grained motion control and synthesis~\cite{burgert2025go,geng2025motion,shin2025motionstream,bruce2024genie,tu2025playerone,he2025matrix,li2025hunyuan,feng2024stratified}. We adopt a similar approach: representing actions as 2D spatial maps in the same domain as the video, providing pixel-level guidance for where hands should appear. Specifically, at frame $i$, we project 3D keypoints $P_i$, which includes human skeleton and hand keypoints (see Fig.~\ref{fig:action_rep}), onto the image plane to obtain per timestep action:
\begin{equation}
\label{action_proj}
A_i = \Pi(\zeta_i, P_i, K),
\end{equation}
where $\Pi(\cdot)$ is the 3D-to-2D camera projection. 
We rasterize these projected poses into 2D action maps $\hat{A}\in\mathbb{R}^{3\times L\times H\times W}$ at full video resolution, with background masked out (Fig.~\ref{fig:action_rep}). Keypoints outside the camera frustum, \textit{i.e.}, the human field of view, are discarded to prevent scene information from leaking into the action signal at inference time. \\
\textbf{Joint Action-Environment Modeling.} The central question is: how should actions condition video generation? We find that naively using action as condition signals does not lead to good results. Our solution is to denoise actions and video jointly. 
Following the same spirit of VideoJAM~\cite{chefer2025videojam}, we encode action maps into latents $a = \mathcal{E}(\hat{A})\in\mathbb{R}^{C\times L'\times H'\times W'}$ and concatenate them with camera-conditioned environment latents along the temporal dimension to model the joint action-environment distribution $p_{t}([x^I, x+z, a]\mid c)$, where $[x^I, x+z, a]\in\mathbb{R}^{C\times (2L'+1)\times H'\times W'}$. During training, we add Gaussian noise to both video latents $x$ and action $a$, while the input image latent $x^{I}$ remains clean. We denote $[x^I, x+z, a]$ as $[\hat{x}, a]$ for simplicity. The diffusion model learns to denoise both video latents and actions by minimizing:
\begin{equation}
\mathcal{L} =
\mathbb{E}_{t, q([\hat{x}_1, a_1]), p_t([\hat{x},a]\mid c)}
\left[
\left\| u_{\theta,t}([\hat{x}, a] \mid c) - u_t([\hat{x}, a] \mid c) \right\|_2^2
\right],
\end{equation}
where $u_{\theta,t}$ and $u_{t}$ are the predicted and target velocities, and $c=\{y, z\}$ includes text and camera conditions. \\
\textbf{Modified Guidance.} Standard CFG assumes the condition $c$ is independent of the denoised output. In our case, the action $a$ is itself denoised by the model and it is not independent from the model output anymore. Inspired by Inner Guidance in~\cite{chefer2025videojam}, we modify the sampling distribution to account for both independent conditions $c$ and the non-independent action $a$ as follows:

% \begin{equation}
\begin{equation}
\begin{split}
    & \tilde{p}_{\theta'}([\hat{x}, a]|c) \\
    & \propto p_{\theta'}([\hat{x}, a]|c) p_{\theta'}(c|[\hat{x}, a])^{w_1} p_{\theta'}(a|x, c)^{w_2} \\
    & \propto p_{\theta'}([\hat{x}, a]|c) \left( \frac{p_{\theta'}([\hat{x}, a], c)}{p_{\theta'}([\hat{x}, a])} \right)^{w_1} \left( \frac{p_{\theta'}([\hat{x}, a], c)}{p_{\theta'}(\hat{x}, c)} \right)^{w_2} \\
    &\propto p_{\theta'}([\hat{x}, a]|c) \left( \frac{p_{\theta'}([\hat{x}, a]|c)}{p_{\theta'}([\hat{x}, a]\mid \varnothing)} \right)^{w_1} \left( \frac{p_{\theta'}([\hat{x}, a]|c)}{p_{\theta'}(\hat{x}, \varnothing |c)} \right)^{w_2},
\end{split}
\end{equation}
by taking log derivative on both sides, the corresponding inference guidance is modified as:

\begin{equation} 
\label{inner}
\begin{split} 
& \tilde{u}_{\theta,t}([\hat{x}, a], c) \\
&= (1 + w_1 + w_2) \cdot u_{\theta,t}([\hat{x}, a], c) \\ &\quad - w_1 \cdot u_{\theta,t}([\hat{x}, a]\mid \varnothing) - w_2 \cdot u_{\theta,t}([\hat{x}, \varnothing]\mid c). 
\end{split} 
\end{equation}
After the denoising process with inference guidance in Eq.~\ref{inner}, we generate the final video by discarding the first frame and the latter half of the frames—corresponding to the recovered $x^I$ and the recovered action latents $a$, respectively—from the denoised latent sequence, and then decoding the remaining latents, as shown in Fig.~\ref{fig:pipeline}.

\begin{figure}[t]
    \centering 
    % First Subfigure
    \begin{subfigure}[b]{\linewidth}
        \centering
        % use \textwidth instead of \linewidth for the subfigure width to be safe
        \includegraphics[width=1.0\linewidth]{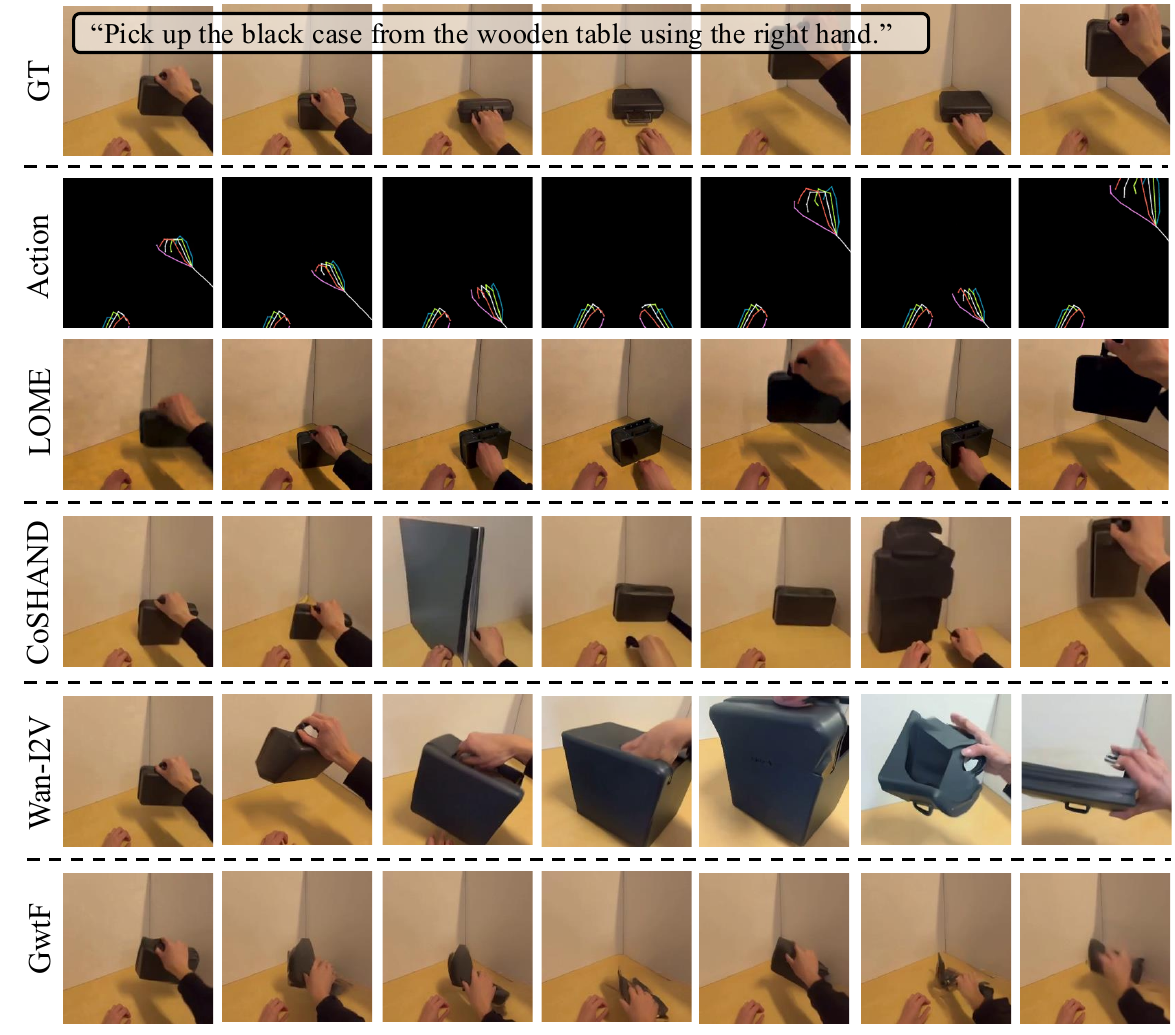}
        \vspace{-6mm}
        \caption{}
        \label{fig:qual1}
    \end{subfigure}
    
    % Leave an empty line here to ensure they stack vertically
    
    % Second Subfigure
    % \begin{subfigure}[b]{\linewidth}
    %     \centering
    %     \includegraphics[width=1.0\linewidth]{figures/qualitative2.pdf}
    %     \vspace{-4mm}
    %     \caption{}
    %     \label{fig:qual3}
    % \end{subfigure}
   
    % Leave an empty line here
    
    \begin{subfigure}[c]{\linewidth}
        \centering
        \includegraphics[width=1\linewidth]{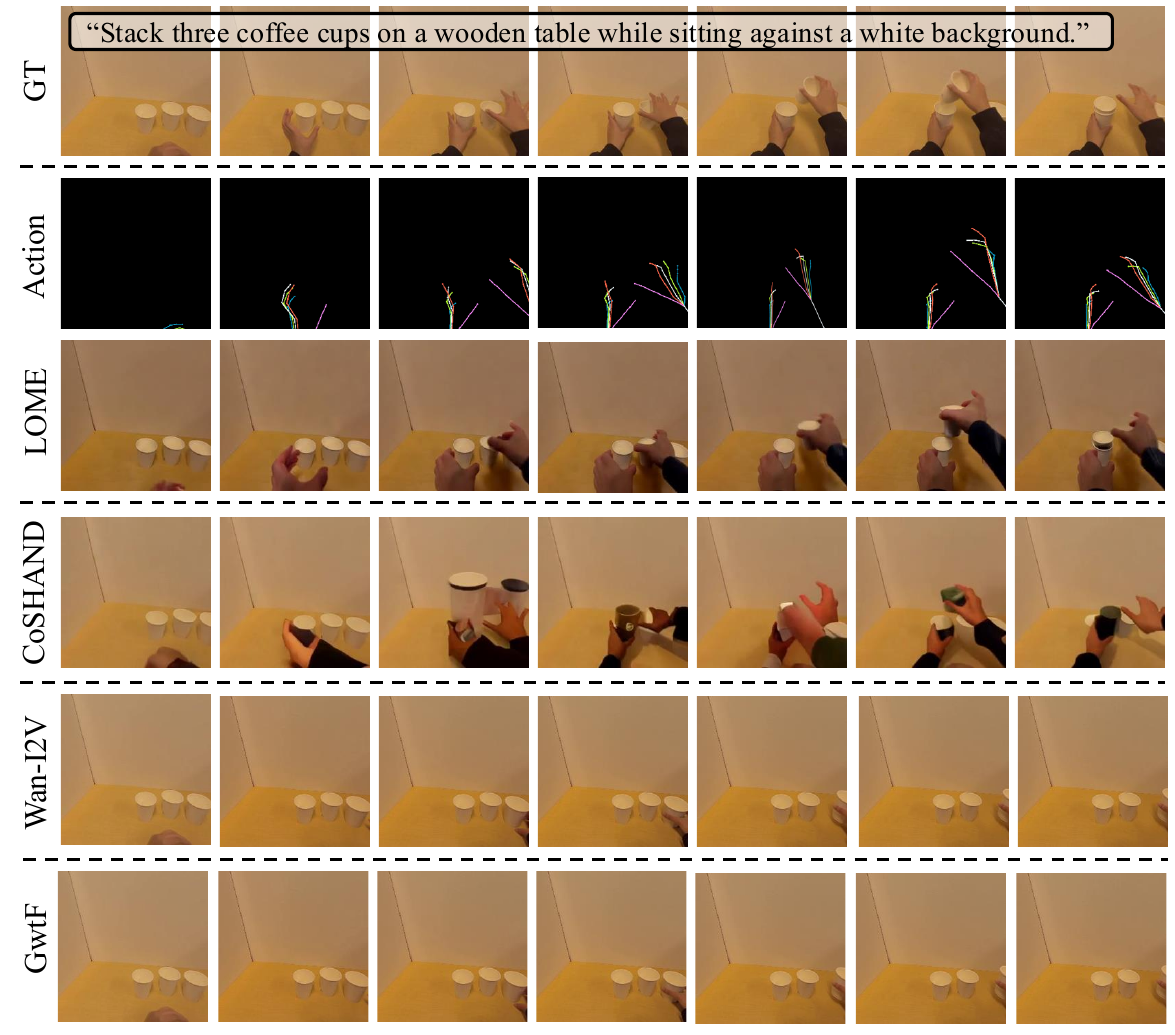}
        \vspace{-6mm}
        \caption{}
        \label{fig:qual2}
    \end{subfigure}
     
    \vspace{-3mm}
    % Main Caption for the whole figure
    \caption{\textbf{Qualitative action-following comparison across tasks.} We compare LOME (ours) with CoSHAND, Wan-I2V and Go-with-Flow (GwtF) on diverse human-object manipulations. “Action” denotes our 2D action maps; CoSHAND uses its own hand masks; Wan-I2V uses no action condition; GwtF uses GT optical flow as action condition. Text prompts are overlaid on the ground-truth (GT) frames.}
    \label{fig:qualitative_comparison}
    \vspace{-6mm}
\end{figure}

\begin{figure}[t]
    \centering
    % First image
    \begin{minipage}[t]{1.0\linewidth} % Adjust width as needed
        \centering
        \includegraphics[width=\linewidth]{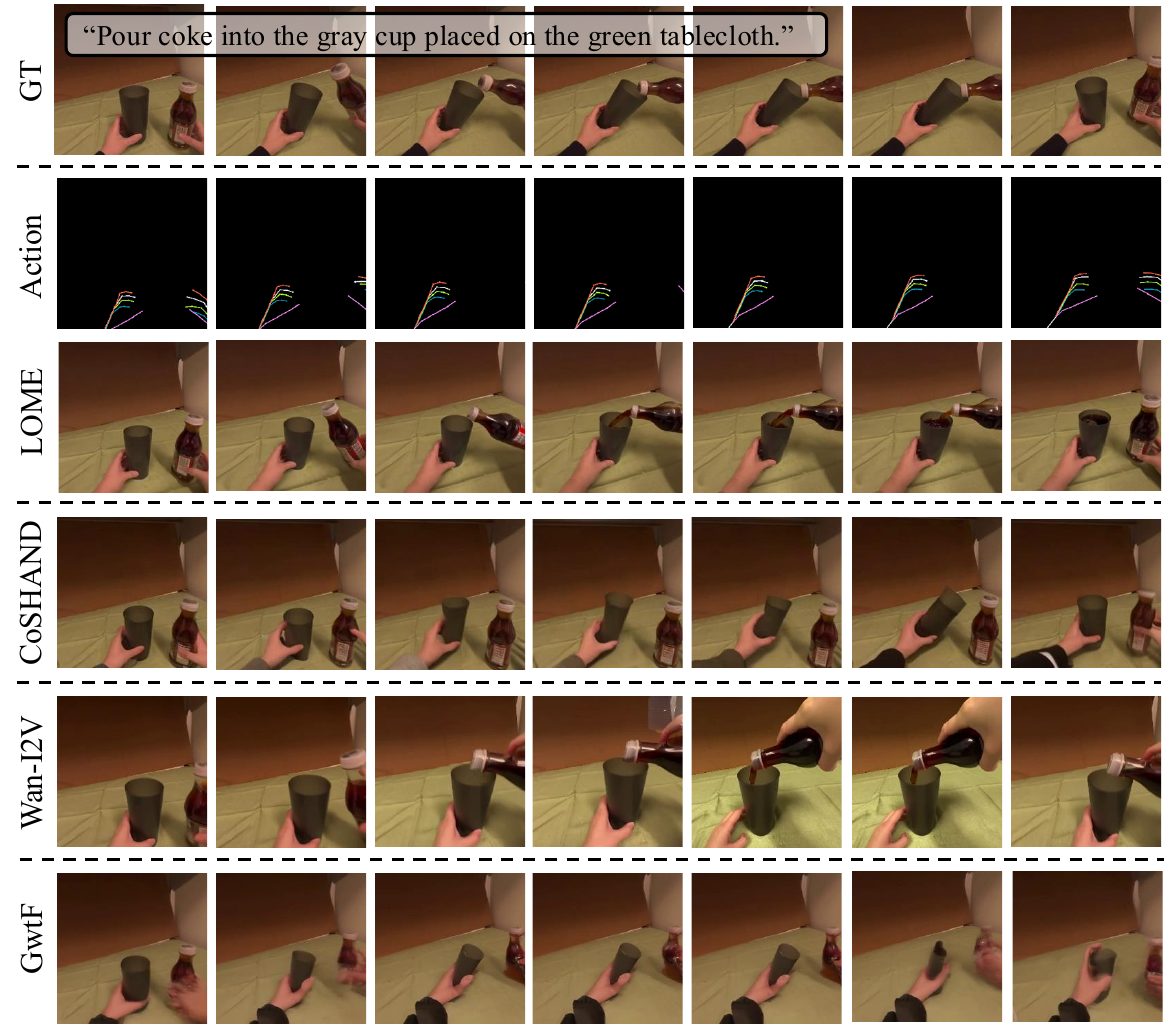} 
        % fig/tsne_2_crop (1).pdf
    \end{minipage}

    \vspace{-2mm}
    \caption{
     \textbf{Pouring example.} We compare LOME (ours), CoSHAND,  Wan-I2V and Go-with-Flow (GwtF) on a ``pouring liquid'' task. 
     % Only LOME successfully hallucinates plausible liquid dynamics with a progressively increasing liquid level over time that aligns with the text prompt. Text prompt is attached to the GT video frames.
     Only LOME produces coherent liquid dynamics with a steadily increasing liquid level consistent with the text instruction. The prompt is overlaid on the GT frames.
     }
     \vspace{-2mm}
    \label{fig:pour_qualitative}
\end{figure}

\section{Experiments}
\subsection{Implementation Details} 
\begin{figure}[th!]
    \centering
    % First image
    \begin{minipage}[t]{1.0\linewidth} % Adjust width as needed
        \centering
        \includegraphics[width=\linewidth]{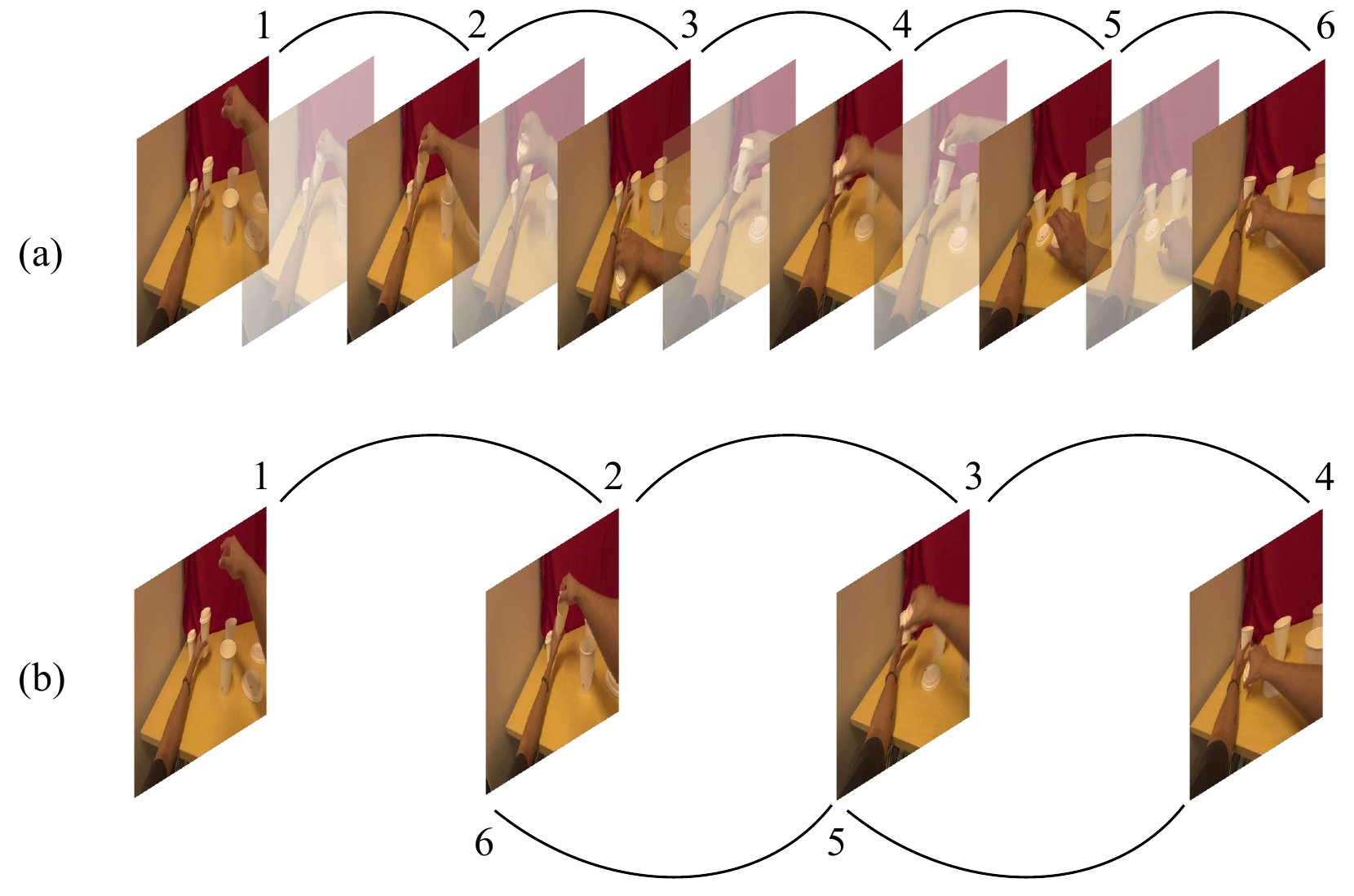} 
        % fig/tsne_2_crop (1).pdf
    \end{minipage}

    \vspace{-2mm}
    \caption{
       \textbf{Temporal resampling to align text and motion.} 
       % To align the text label with the corresponding motion sequence shown in the video, we propose to resample the varying number of video frames into a consistent number, $i.e.$ 6. As illustrated in (a), videos with more than 6 frames are downsampled, while (b) shows the upsampling process for videos with fewer than 6 frames.
       We propose to resample clips with the varying number of frames to a fixed length ($i.e.$ 6 frames). (a) Longer clips are uniformly downsampled while preserving the first and last frames. (b) Shorter clips are upsampled by back-and-forth resampling to reach the target length.
    }
    \vspace{-3mm}
    \label{fig:frame_sampling}
\end{figure}

\textbf{Architecture and Training/Inference Strategy.} 
We build on Wan2.1-VACE-14B~\cite{wan2025wan,jiang2025vace}, an open-source model with state-of-the-art performance in conditional video generation. We use its pretrained VAE encoder $\mathcal{E}$, decoder $\mathcal{D}$, and umT5 text encoder without modification.
To preserve the rich motion priors of the base model as much as possible, we freeze the diffusion DiT blocks and apply low-rank adaptation (LoRA) fine-tuning (rank 128) only to the VACE module. In Fig.~\ref{fig:pipeline}, the architecture of Wan2.1-VACE is simplified by combining the VACE block and the diffusion DiT into a single module. For the action condition, we extract 2D action maps by Eq.~\ref{action_proj} for each video. The camera adapter $\mathcal{C}(\cdot)$ consists of a single 2D convolutional layer and a residual block. We train our model for 2,000 steps on 32 NVIDIA A100-80GB GPUs with a global batch size of 32 (one video per GPU), using a learning rate of $1\times10^{-5}$ and a sample resolution of $832\times480$. The number of frames per video is set to 49 in training. We adopt the settings $w_1=5$ and $w_2=3$ from VideoJAM~\cite{chefer2025videojam}, while extending the inference sequence length to 81 frames to match Wan-I2V for comparison.\\
% In Fig.~\ref{fig:pipeline}, the architecture of Wan2.1-VACE is simplified by combining the VACE block and the diffusion DiT into a single module. In our implementation, we adopt the pre-trained Wan2.1 encoder $\mathcal{E}(\cdot)$, decoder $\mathcal{D}(\cdot)$, and umT5 as the text encoder. For the action condition, we extract 2D action maps by Eq.~\ref{action_proj} for each video. The camera adapter $\mathcal{C}(\cdot)$ consists of a single 2D convolutional layer and a residual block. We train our model for 2,000 steps on 32 NVIDIA A100-80GB GPUs with a global batch size of 32 (one video per GPU), using a learning rate of $1\times10^{-5}$ and a sample resolution of $832\times480$. The number of frames per video is set to 49. We adopt the settings $w_1=5$ and $w_2=3$ from VideoJAM~\cite{chefer2025videojam}, while extending the inference sequence length to 81 frames to match Wan-I2V for comparison.\\
\textbf{Learning Complete Manipulation.} 
We resample all input video frames, action maps, and corresponding camera parameters to ensure equal-length clips that capture the start and the end of complete human–object manipulations, thereby aligning the videos with the semantic content of the corresponding text annotations. Specifically, when the input video contains more frames than required by the model, we downsample it by uniformly selecting frames while always preserving the first and last frames, as shown in Fig.~\ref{fig:frame_sampling} (a). When the input video contains fewer frames than required, we temporally resample the sequence in a back-and-forth manner until the target video length is reached, as in Fig.~\ref{fig:frame_sampling} (b). 

\subsection{Dataset} 
\textbf{EgoDex.} EgoDex~\cite{hoque2025egodex} is an egocentric human video dataset comprising 338{,}234 short videos with a resolution of $1920 \times 1080$ and approximately 800 hours of footage. It captures a diverse range of human–object manipulation scenarios using Apple Vision Pro and provides detailed 3D human pose annotations, including the hands, arms, spine, and neck, as shown in Fig.~\ref{action_proj} (a). Per-timestep camera extrinsics $\zeta_i$ are estimated via on-device SLAM, while the camera intrinsics $\mathbf{K}$ are assumed to be known and constant across all videos. We train LOME with videos sampled from the union of 5 training sets of EgoDex and reserve 1 test set (with no overlap with the training data) for inference. \\
\textbf{In-the-wild Video Captures.} To showcase the generalization ability, we additionally record
10 egocentric videos in our labs of hand-object manipulations with daily objects. These videos include novel objects and environments.

\subsection{Baselines and Metrics} 
\textbf{CoSHAND.} The most related work is CoSHAND~\cite{sudhakar2024controlling}, an image diffusion model conditioned on human hand actions. Rather than generating videos of continuous human–object interactions, CoSHAND synthesizes target images conditioned on a source image and target hand masks, with the generated images following the specified hand poses.  \\
\textbf{Wan2.1-I2V-14B.} This is a text and image-conditioned video generative model and we refer to this model as ``Wan-I2V"~\cite{wan2025wan} for brevity. Despite not supporting action-conditioned generation, pretrained Wan-I2V is compared with our method in terms of visual quality and text-following.  \\ 
\textbf{Go-with-the-Flow.} Video generative models with spatial control have proven effective for motion synthesis. One of representative works Go-with-the-Flow~\cite{burgert2025go} conditions video generation on optical flow instead of human actions. In our implementation, we directly extract optical flow from GT videos and use it to warp the input image. We refer to this model as ``GwtF" for brevity.\\
\textbf{Metrics.} We evaluate action-following performance using \\PCK@20 ~\cite{goswami2025world}, which measures the percentage of keypoints that fall within a 20-pixel radius of their ground-truth locations. Specifically, we use MediaPipe~\cite{mediapipe} to extract 21 hand keypoints and compute PCK@20 between the keypoints detected in the ground-truth videos and those from our generated videos. We exclude frames without detected hands from the evaluation and report the ratio of frames with no hand overlap or detection failure over all inference frames (81). Beyond action-following, we evaluate motion consistency using FVD, tLPIPS, and CLIP-I, where lower tLPIPS indicates reduced temporal jitter and flicker. CLIP-I measures how well the semantic identity of objects is preserved compared to the real videos, while CLIP-S evaluates text–video alignment. We further assess visual quality by computing LPIPS, SSIM, and PSNR between corresponding frames of the ground-truth (GT) videos and the generated videos. Following~\cite{wang2025physctrl,bansal2024videophy,meng2024towards}, we evaluate Semantic Adherence (SA) and Physical commonsense (PC) to measure physics realism of generated videos in a 5-like score by GPT.
\begin{table}[h]
\centering
% \vspace{-3mm}
\caption{\textbf{Quantitative comparisons on motion consistency and action-following.} We report FVD, CLIP-I, CLIP-S, tLPIPS, and PCK@20 across methods (arrows indicate whether higher or lower is better).}
\vspace{-2mm}
\resizebox{1.0\linewidth}{!}{
\begin{tabular}{lccccccc}
\toprule
Method & FVD$\downarrow$ & CLIP-I$\uparrow$ & CLIP-S$\uparrow$ & tLPIPS$\downarrow$ & PCK@20$\uparrow$  \\
\midrule
CoSHAND & 59.83 & 0.846 & 0.267  & 0.171 & 51.33  \\
Wan-I2V & 68.77 & 0.884 & 0.294  &  0.057 & 10.71 \\
GwtF & 40.33 & 0.878 & 0.256 &  0.071 & 47.06  \\
LOME (ours) & \textbf{39.58} & \textbf{0.903} & 0.284  & \textbf{0.051} & \textbf{66.85} \\
\bottomrule
\end{tabular}
}
\label{tab: comp2}
\vspace{-2mm}
\end{table}
\subsection{Results and Analysis}
\textbf{Qualitative Evaluation.} We conduct qualitative comparisons between LOME, CoSHAND, Wan-I2V and GwtF across different tasks as shown in Fig.~\ref{fig:qualitative_comparison}. In each task, we sample 7 out of 81 frames from the corresponding video for visualization. All methods are conditioned on the first frame and a text prompt of the ground-truth video. Except for Wan-I2V, which does not use action conditions, CoSHAND requires hand masks segmented from the ground-truth video frames as condition signals, GtwF warps the input image by optical flow extracted from the ground-truth video, whereas our method uses action maps from 3D human pose estimations. Since CoSHAND generates images at a resolution of $256 \times 256$, we resize generative results of Wan-I2V, LOME and our action maps from $832 \times 480$ to $256 \times 256$ for better visual alignment with CoSHAND, while GT videos are resized from  $1920 \times 1080$ and GwtF videos are resized from $854 \times 480$. Videos at their original resolutions are included in the supplementary material. 
Qualitative comparisons demonstrate that LOME significantly outperforms CoSHAND, Wan-I2V and GwtF in terms of both object and hand motion consistency, and produces more realistic hand–object interactions. Moreover, CoSHAND struggles to interact with the correct object when multiple objects are present, while Wan-I2V fails to generate complete manipulation sequences as described by the text prompts.  

\begin{figure*}[t]
    \centering
    % First image
    \begin{minipage}[t]{1.0\linewidth} % Adjust width as needed
        \centering
        \includegraphics[width=\linewidth]{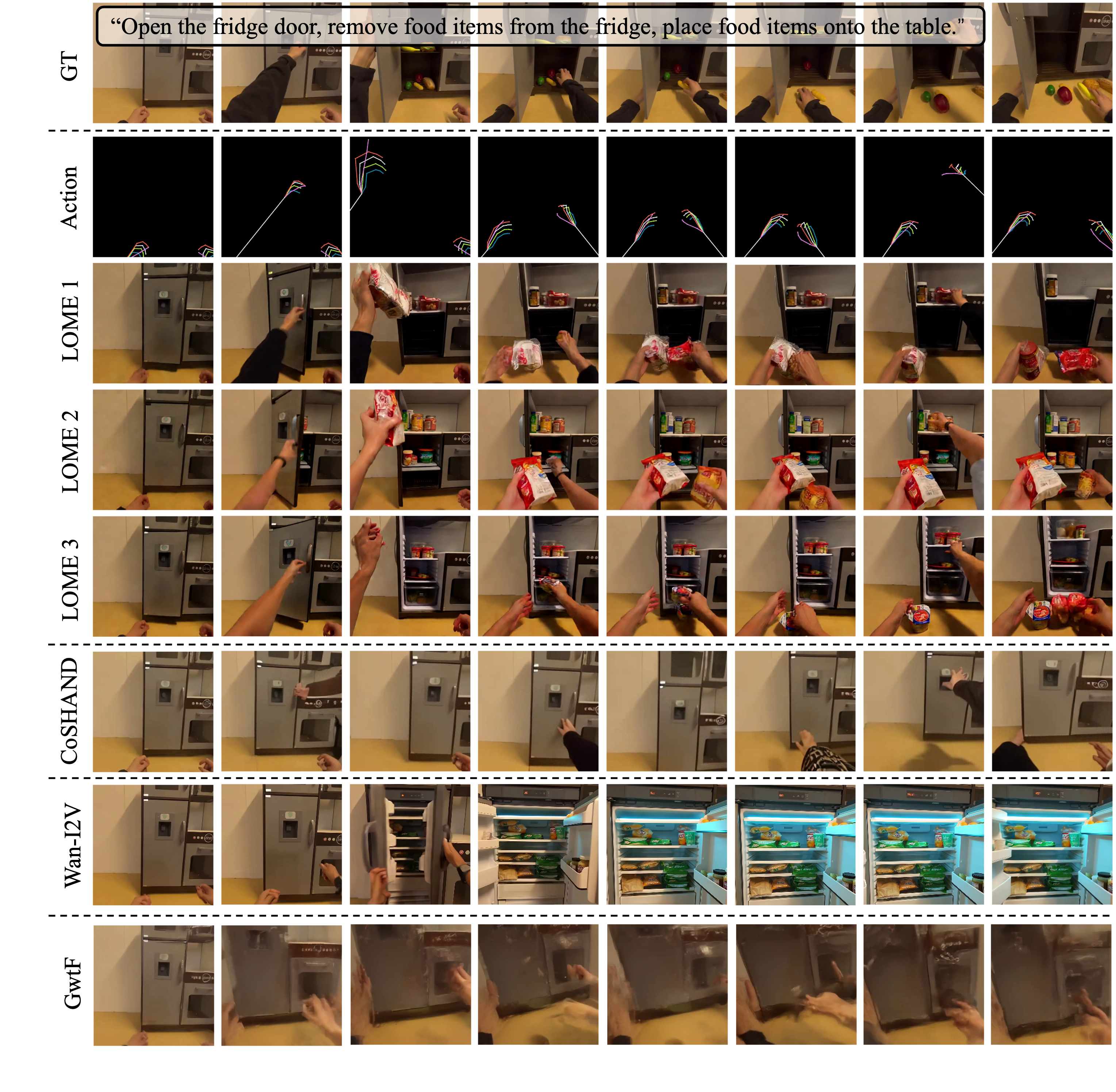} 
        % fig/tsne_2_crop (1).pdf
    \end{minipage}
   
    % \begin{minipage}[t]{1.0\linewidth} % Adjust width as needed
    %     \centering
    %      \vspace{-3mm}
    %     \includegraphics[width=\linewidth]{figures/diversity_sup.pdf} 
    %     % fig/tsne_2_crop (1).pdf
    % \end{minipage}
    \vspace{-5mm}
    \caption{
    \textbf{Occluded-object manipulation and output diversity.}
     We compare LOME (ours), CoSHAND, Wan-I2V and GwtF on a challenging task where some of the objects to be manipulated are not visible in the input image (e.g., behind the fridge door). Among the three methods, only LOME produces plausible human–object interactions in this setting. LOME 1-3 denote three stochastic inference runs under identical conditions, illustrating output diversity. Text prompt is overlaid on the GT video frames.}
   
    \label{fig:diverse}
\end{figure*}

We further compare LOME with other methods on a challenging case shown in Fig.~\ref{fig:pour_qualitative}, where the coke bottle in the input image has its cap tightly fastened. Among all methods, only LOME is able to generate videos that closely follow the text prompt, realistically depicting liquid flowing from the bottle into the cup. Moreover, despite slight temporal variations in the action conditions, the amount of liquid in the cup increases progressively over time, reflecting coherent physical consequences of the pouring action. In contrast, CoSHAND and GwtF fail to produce meaningful hand–object interactions, while Wan-I2V does not generate a complete pouring sequence and the cup does not become filled as the video progresses. We conduct a comprehensive user study on 10 examples generated by LOME and other compared methods. For each example, users are asked to vote for the method that best demonstrates text adherence, action adherence, motion consistency, and visual quality. The results are shown in Tab.~\ref{tab: user_study}.\\
\textbf{Diversity Evaluation.} We showcase diverse generative results under the same text prompt and input image in Fig.~\ref{fig:diverse}. In this case, some of the objects to be manipulated are behind the fridge door and not visible in the input image. Notably, only LOME can generate plausible and diverse hand–object interaction sequences, whereas Wan-I2V fails to synthesize meaningful hand motions and CoSHAND or GwtF struggles to open the fridge door and hallucinate the objects behind it.\\
\textbf{Quantitative Evaluation.} All quantitative evaluations are conducted at a resolution of $256 \times 256$. As shown in Tab.~\ref{tab: comp1}, our LOME achieves the best physics realism. Although CoSHAND achieves higher PSNR and SSIM, this is primarily because  CoSHAND is an image-based diffusion model, which preserves per-frame texture, lighting, and background details more faithfully than video-based models. In contrast, video-based models are more prone to temporal drift, which can negatively impact pixel-level metrics. Nevertheless, LOME achieves the best motion consistency and action-following performance among the compared methods, as demonstrated in Tab.~\ref{tab: comp2} and Tab.~\ref{tab: user_study}.

\begin{table}[t]
\centering
% \vspace{-4mm}
\caption{\textbf{User study results.} We collect votes
from 30 participants over 10 test samples. LOME receives the highest percentage of votes for best text-following (TF), action-following (AF), motion consistency (MC), and visual quality (VQ).}
\vspace{-3mm}
\resizebox{0.7\linewidth}{!}{
\begin{tabular}{lccccc}
\toprule
Method & TF $\uparrow$ & AF$\uparrow$ & MC$\uparrow$ & VQ$\uparrow$  \\
\midrule
CoSHAND & 0.67 & 0.67  & 0.00  & 2.01 \\
Wan-I2V &  1.34  & 0.00 & 2.01 & 4.02 \\
GwtF &  0.67  & 0.67 & 0.67  & 0.00 \\
LOME (ours) & \textbf{97.32} & \textbf{98.66}  & \textbf{97.32} &  \textbf{93.97} \\
\bottomrule
\end{tabular}
}
\label{tab: user_study}
\vspace{-2mm}
\end{table}

\begin{table}[t]
\centering
% \vspace{-4mm}
    \caption{\textbf{Quantitative comparison on visual quality and physics realism.} For visual quality, we report PSNR, SSIM, and LPIPS between generated and ground-truth videos (arrows indicate whether higher or lower is better). For physics realism, we report Semantic Adherence (SA) and Physical Commonsense (PC).}
\vspace{-3mm}
\resizebox{0.8\linewidth}{!}{
\begin{tabular}{lcccccc}
\toprule
Method & PSNR$\uparrow$ & SSIM$\uparrow$ & LPIPS$\downarrow$ & SA$\uparrow$ & PC $\uparrow$\\
\midrule
CoSHAND & \textbf{19.35} & \textbf{0.731} & 0.366 & 1.4 & 1.1\\
Wan-I2V & 16.53  & 0.624 & 0.489 & 3.3 & 3.6\\
GwtF &  18.93 & 0.704 & 0.435 & 2.6 & 2.1\\
LOME (ours) & 18.88 & 0.719 & \textbf{0.325} & \textbf{4.1} & \textbf{4.0}\\
\bottomrule
\end{tabular}
}
\label{tab: comp1}
\vspace{-3mm}
\end{table}
Current I2V/T2V models depend heavily on detailed text prompts, leaving motion synthesis underconstrained when such descriptions are absent. Our experiments show that baselines like Wan-I2V and CoSHAND exhibit limited controllability and consistency. Conversely, LOME integrates explicit spatial action control into a video-based backbone, significantly enhancing motion realism. We also observe that GwtF struggles to synthesize fine-grained hand motions, even when conditioned on ground truth optical flow. While Wan-I2V outperforms our method in CLIP-S (text alignment), qualitative results indicate that LOME provides a much closer match to the ground truth videos in terms of visual and temporal fidelity.
\subsection{Ablation Study} 
\begin{figure}[t]
    \centering
    % First image
    \begin{minipage}[t]{1.0\linewidth} % Adjust width as needed
        \centering
        \includegraphics[width=\linewidth]{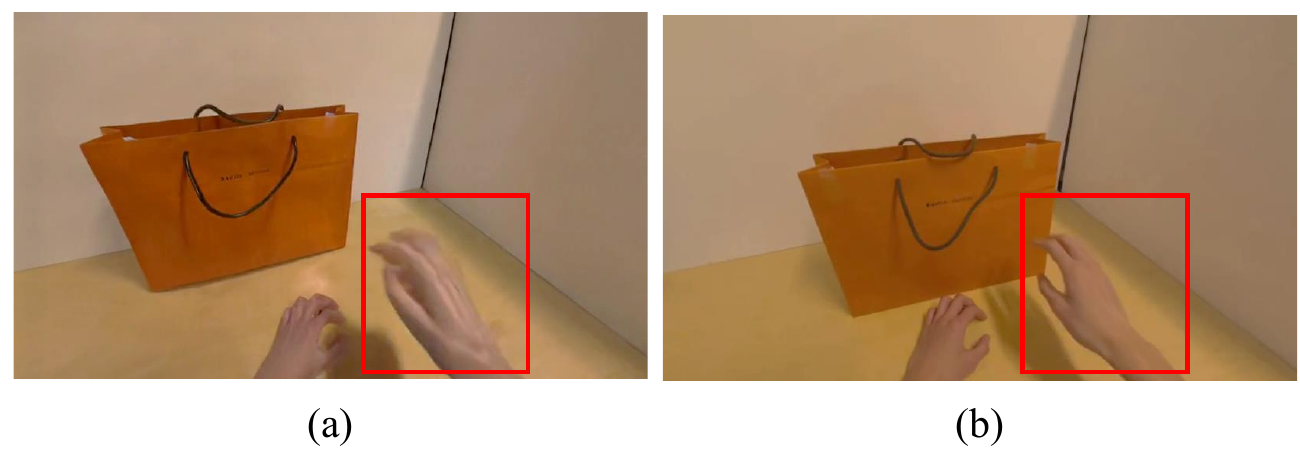} 
        % fig/tsne_2_crop (1).pdf
    \end{minipage}

    \vspace{-2mm}
    \caption{
      \textbf{Ablation on joint action-environment modeling.} (a) LOME without joint action–environment modeling, where no Gaussian noise is added to the action maps during training and the original CFG is applied. (b) LOME with joint action–environment modeling.
    }
   \vspace{-4mm}
    \label{fig:ablation_qual}
\end{figure}

We demonstrate the effectiveness of our joint action–environment modeling and inference guidance through qualitative and quantitative ablation studies, as shown in Fig.~\ref{fig:ablation_qual} and Tab.~\ref{tab: ablation}, respectively. Without joint action-environment modeling, $i.e.$ ``Ours (w/o joint modeling)", LOME exhibits degraded hand realism, motion consistency and action-following performance.

Moreover, our joint modeling via temporal concatenation differs from VideoJAM~\cite{chefer2025videojam}, which employs channel concatenation. By comparing with LOME but with channel concatenation $i.e.$ ``Ours (channel concatenation)'', we find that temporal concatenation (``Ours'') leads to better performance, as the bidirectional attention in diffusion modules enables explicit temporal attention between corresponding frames of the video and action maps. 
\begin{table}[t]
\centering
% \vspace{-2mm}
\caption{\textbf{Ablation on motion consistency and action following.} We evaluate the effects of removing joint modeling, removing camera adapter, and replacing temporal concatenation of action and video latents with channel concatenation. We report FVD, CLIP-I, tLPIPS, and PCK@20.}
\vspace{-2mm}
\resizebox{1.0\linewidth}{!}{
\begin{tabular}{lcccccc}
\toprule
Method  & FVD$\downarrow$ & CLIP-I$\uparrow$ & tLPIPS$\downarrow$ & PCK@20$\uparrow$  \\
\midrule
Ours (w/o joint modeling) & 48.01 & 0.881 & 0.063 & 62.81 \\
Ours (channel concatenation) & 41.23  &  0.900  & 0.055 & 65.77  \\
Ours (w/o camera adapter) & 39.51 & 0.903 & 0.049 & 66.84 \\
Ours & \textbf{39.78} & \textbf{0.903} & \textbf{0.051} & \textbf{66.85}\\
\bottomrule
\end{tabular}
}
\vspace{-2mm}
\label{tab: ablation}
\end{table}

\section{Limitations and Future Works}
\begin{figure}[t]
% \vspace{-2mm}
    \centering
    % First image
    \begin{minipage}[t]{1.0\linewidth} % Adjust width as needed
        \centering
        \includegraphics[width=\linewidth]{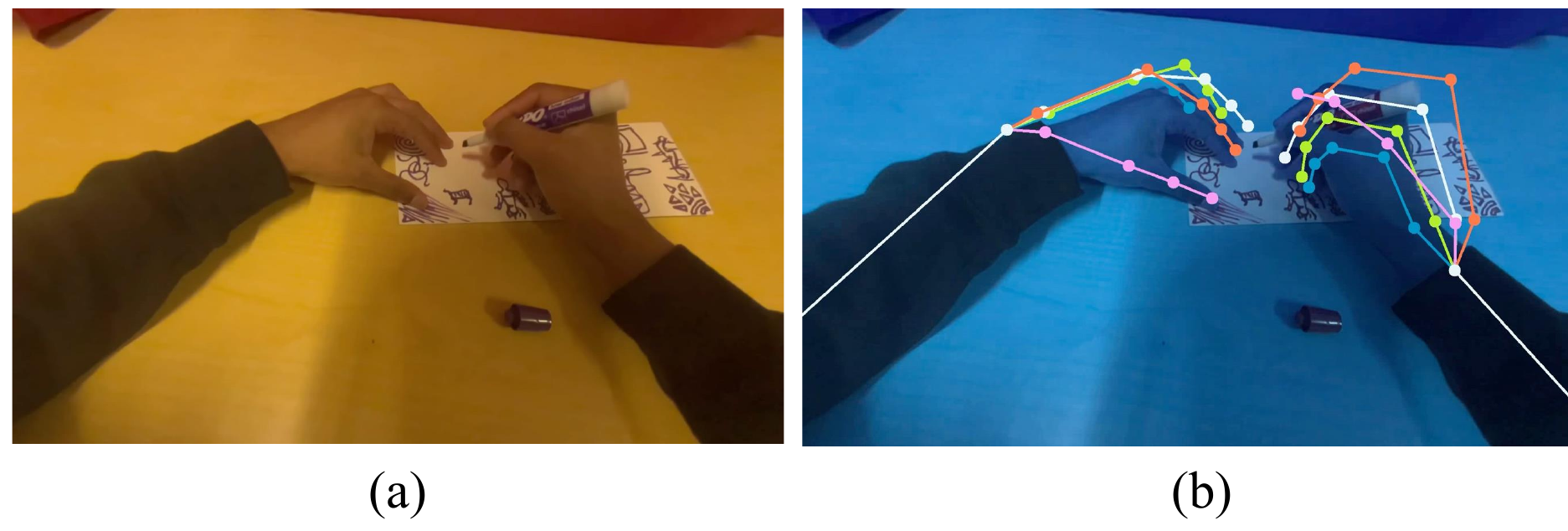} 
        % fig/tsne_2_crop (1).pdf
    \end{minipage}

    \vspace{-3mm}
    \caption{
      \textbf{Examples of Noisy Dataset}. We show an example of projection misalignment due to 3D human pose and camera estimation errors.
      % We visualize misalignments in some cases between the true hand positions and the corresponding action maps that arise from errors in 3D human pose and camera pose estimation. 
      % (a) presents the original image, and (b) shows the overlaid action maps and actual hand positions. 
      (a) Input frame. (b) Overlay of the rasterized action map and the true hand positions, highlighting spatial offsets between the conditioning signal and the observed hands.
    }
   
    \label{fig:projection_error}
     \vspace{-5mm}
\end{figure}

\begin{figure}[t]
    \centering
    % First image
    \begin{minipage}[t]{1.0\linewidth} % Adjust width as needed
        \centering
        \includegraphics[width=\linewidth]{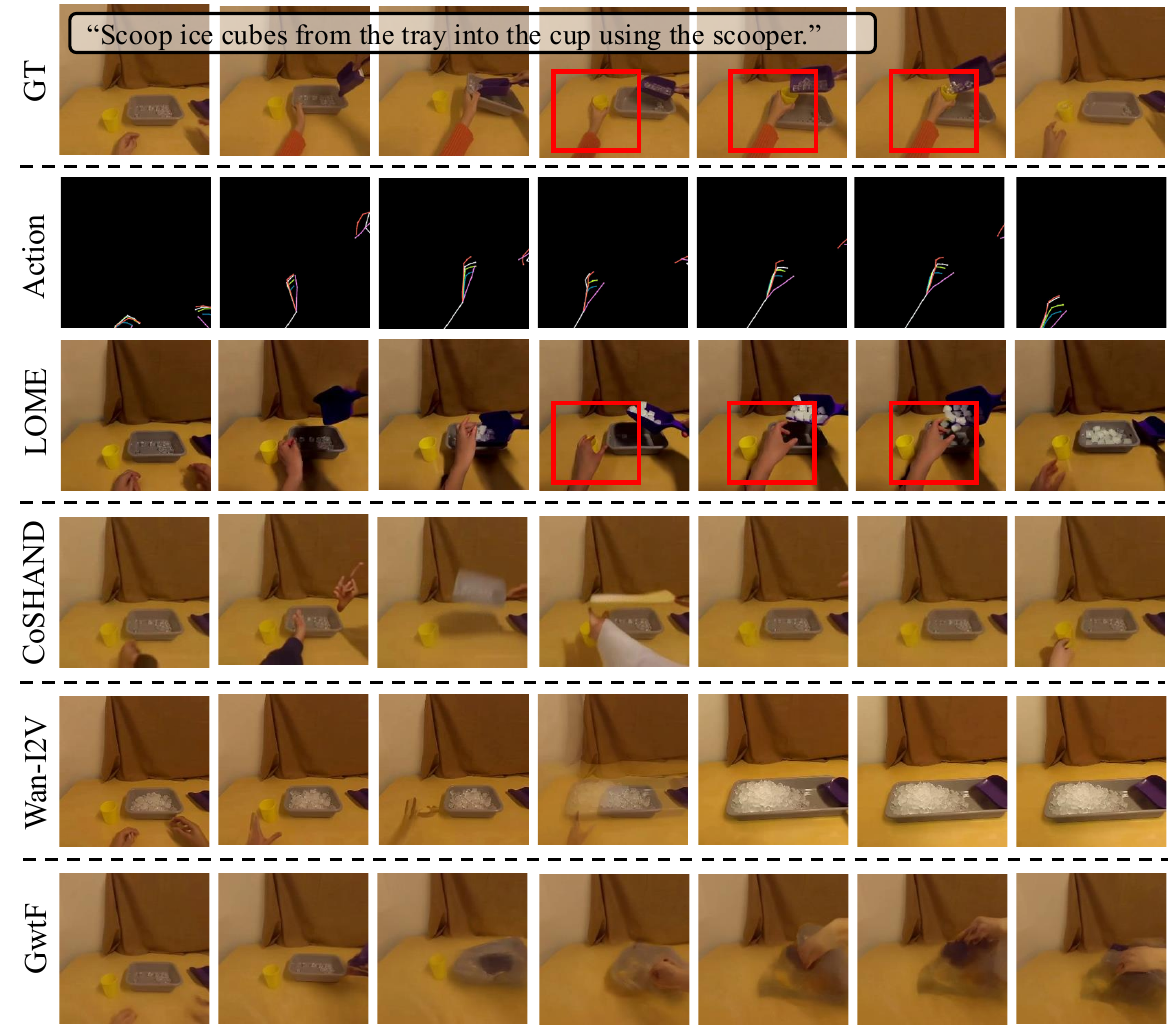} 
        % fig/tsne_2_crop (1).pdf
    \end{minipage}

    \vspace{-2mm}
    \caption{
     \textbf{Failure case with multi-object interaction.} 
     % We present a failure case of LOME that requires simultaneous interaction with multiple objects. In this example, LOME struggles to correctly grasp the cup, and the ice cubes fall into the tray instead of the cup (highlighted in red). Text prompt is attached to the GT video frames.
     In this example, LOME struggles to coordinate simultaneous interactions: it fails to correctly grasp the cup, causing ice cubes to fall into the tray rather than the cup (highlighted in red). The prompt is overlaid on the GT frames.
    }
   
    \label{fig:failure}
     % \vspace{-4mm}
\end{figure}

As shown in Fig.~\ref{fig:projection_error}, 3D human pose and camera pose estimation from headsets are imperfect, which can lead to projection misalignments between action maps and hand poses, resulting in lower-than-expected PCK@20 scores. We also present a failure case in Fig.~\ref{fig:failure}. While LOME significantly outperforms the baselines, it still struggles with coordinating interactions among multiple objects. Specifically, the model fails to grasp and lift the yellow cup, causing the ice cubes to fall into the tray rather than the target cup specified in the text prompt. As a direction for future work, we plan to leverage distillation techniques to enable autoregressive inference and improve efficiency.

\section{Conclusion}
In this paper, we present LOME, an egocentric world model that adapts video diffusion models for learning general human–object manipulation. By jointly modeling human actions and environment, LOME synthesizes realistic, contact-rich manipulation videos that exhibit accurate action-following and high visual fidelity—all without relying on 3D/4D reconstruction or parametric model fitting. Moreover, LOME demonstrates diverse generative results given same conditions. Our results demonstrate the potential of LOME for photorealistic simulation of real-world human-object manipulation.

% \newpage
\bibliographystyle{ACM-Reference-Format}
\bibliography{sections/reference}

\begin{figure*}[t]
    \centering
    % First image
    \begin{minipage}[t]{1.0\linewidth} % Adjust width as needed
        \centering
        \includegraphics[width=\linewidth]{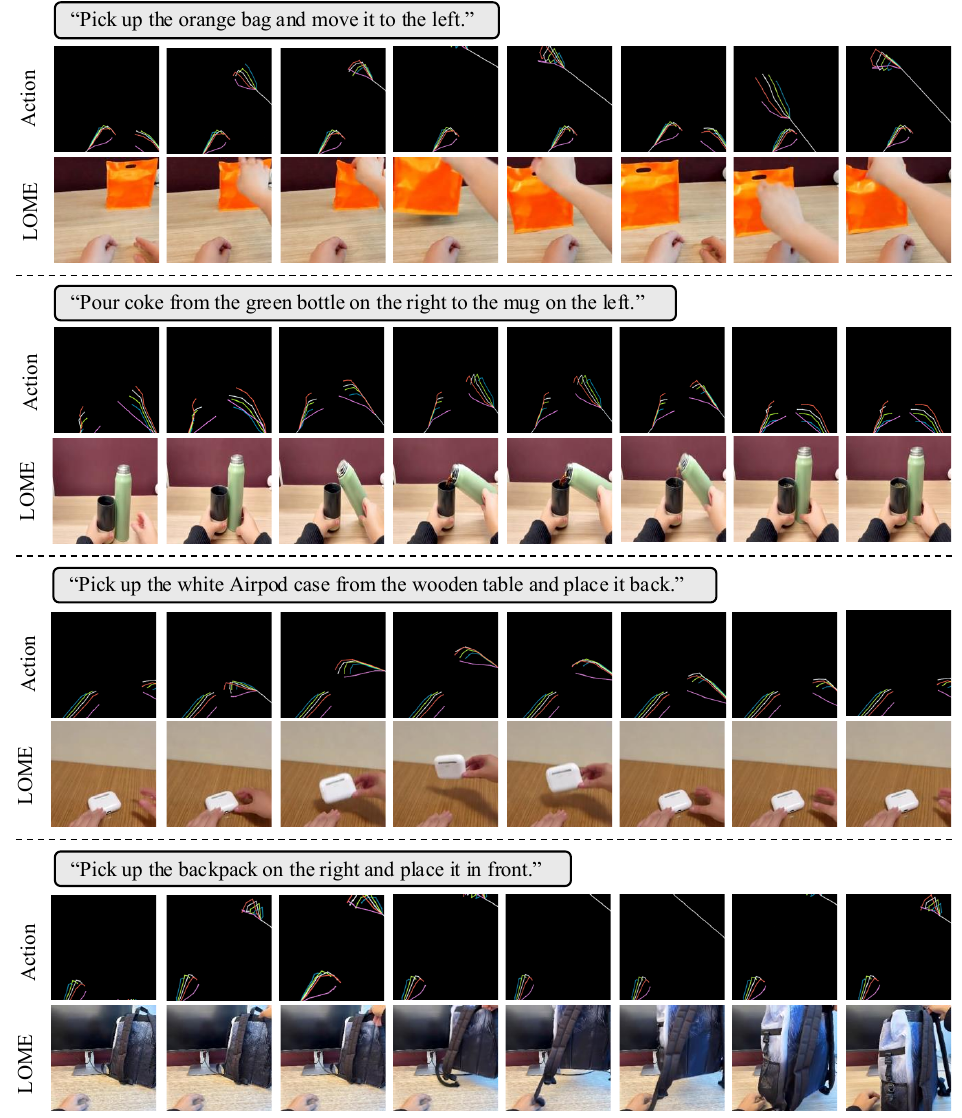} 
        % fig/tsne_2_crop (1).pdf
    \end{minipage}

    % \vspace{-3mm}
    \caption{
     \textbf{In-the-wild lab captures.} 
     % We showcase LOME in in-the-wild scenarios, where the environments and objects are from our lab.
     We showcase LOME on real-world egocentric scenes recorded in our lab with novel objects and environments, demonstrating generalization beyond the training data.}
   
    \label{fig:in_the_wild}
\end{figure*}

\clearpage
\newpage

\section{Appendix}
\begin{figure}[h]
    \centering
    % First image
    \vspace{-6mm}
    \begin{minipage}[t]{1.0\linewidth} % Adjust width as needed
        \centering
        \includegraphics[width=\linewidth]{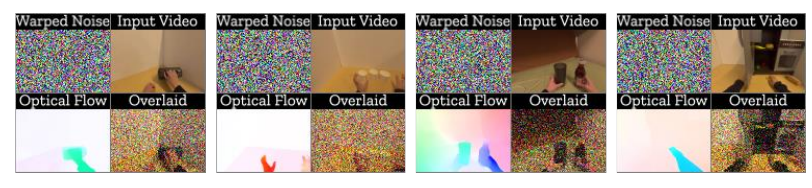} 
    \end{minipage}
    \vspace{-4mm}
    \caption{
     \textbf{Optical flow condition of GwtF.} We visualize the optical flow condition used by GwtF to warp the Gaussian noise corresponding to the input image. Optical flow is extracted from the ``Input Video'' $i.e.$ GT video.}
    \vspace{-4mm}
    \label{fig:gwtf}
\end{figure}

\begin{figure}[h]
    \centering 
    % First Subfigure
    \begin{subfigure}[b]{\linewidth}
        \centering
        % use \textwidth instead of \linewidth for the subfigure width to be safe
        \includegraphics[width=1.0\linewidth]{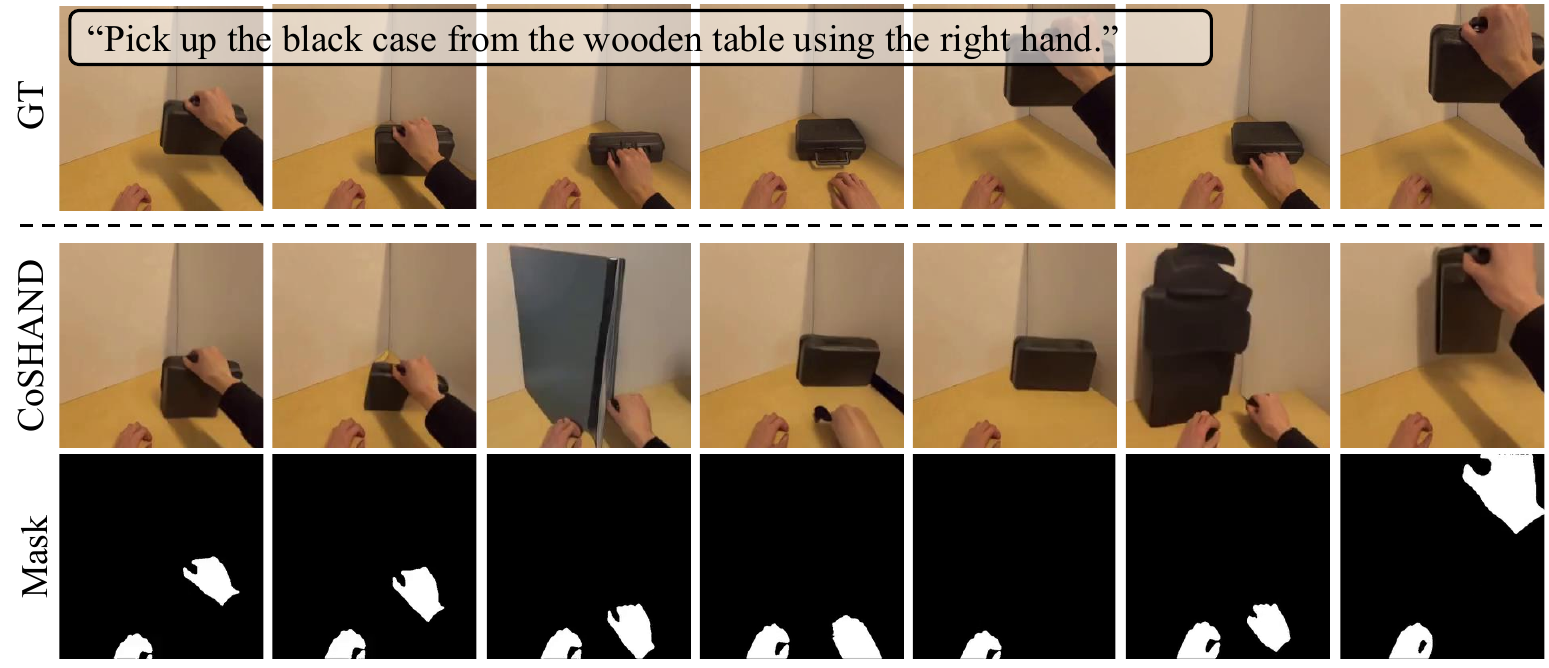}
        \vspace{-6mm}
        \caption{}
        \label{fig:qual1}
    \end{subfigure}
    
    % Leave an empty line here to ensure they stack vertically
    
    % Second Subfigure
    \begin{subfigure}[]{\linewidth}
        \centering
        \includegraphics[width=1.0\linewidth]{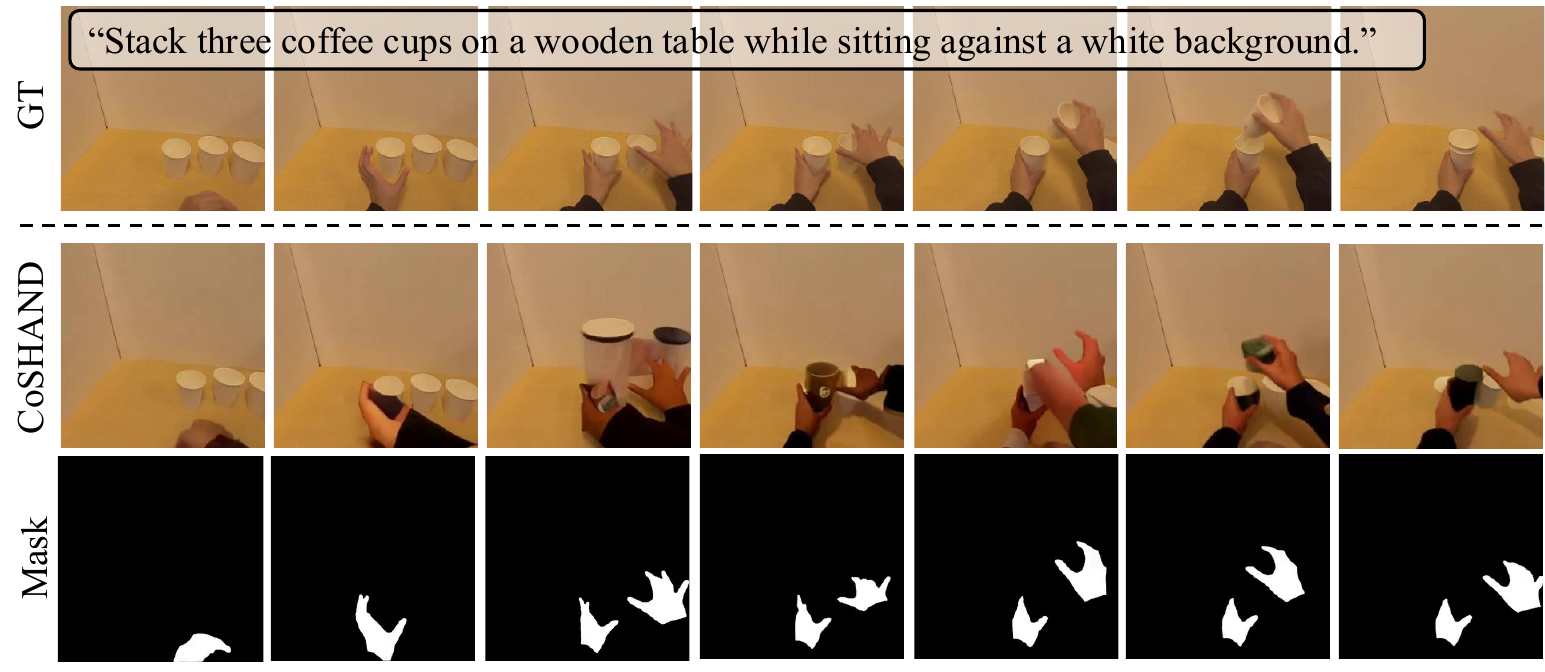}
        \vspace{-6mm}
        \caption{}
        \label{fig:qual3}
    \end{subfigure}
   
    % Leave an empty line here
    
    % \begin{subfigure}[c]{\linewidth}
    %     \centering
    %     \includegraphics[width=1\linewidth]{figures/sup_fig_3.pdf}
    %     \vspace{-5mm}
    %     \caption{}
    %     \label{fig:qual2}
    % \end{subfigure}
     
    \vspace{-2mm}
    % Main Caption for the whole figure
    \caption{\textbf{Hand mask visualization of CoSHAND.} We visualize the hand mask of CoSHAND on the same examples as in Fig. 4 of the main paper.}
    \label{fig:hand_mask}
    \vspace{-4mm}
\end{figure}

\textbf{Optical Flow Condition of GwtF.} We follow the official code of GwtF in implementation. As shown in Fig.~\ref{fig:gwtf}, GwtF uses optical flow to warp the Gaussian noise of the input image, producing noise for subsequent frames that serves as the condition for video generation. Ideally, only hand motion ($i.e.$ optical flow) should be warped as object motion should emerge as a consequence of human action, rather than being determined in advance. Nevertheless, we directly use optical flow extracted from GT videos to showcase the best possible performance of GwtF in all comparisons. \\
\textbf{Action Condition of CoSHAND.} We follow the official code of CoSHAND in implementation, CoSHAND applies the off-the-shelf segmentation algorithm SegAny~\cite{kirillov2023segment} to each frame of the GT video to obtain hand masks as the action condition. As shown in Fig.~\ref{fig:hand_mask}, these hand masks are more precise than our action maps, since SegAny operates directly on GT videos, whereas our action maps are obtained without using any GT video information, relying instead on 2D projections of estimated 3D human poses. However, per-frame segmentation is not robust to extreme hand poses and can lose track of the hand (e.g. hand mask shown in Fig.~\ref{fig:hand_mask}(c)). Despite using less accurate hand actions, LOME consistently produces more reasonable hand–object interactions than CoSHAND. \\
% \textbf{More Qualitative Results.} Please refer to Fig.~\ref{fig:more_qualitative_comparison} and supplementary video for more qualitative comparisons.
\begin{figure}[t]
    \centering
    % First image
     \vspace{-2mm}
    \begin{minipage}[t]{1.0\linewidth} % Adjust width as needed
        \centering
        \includegraphics[width=\linewidth]{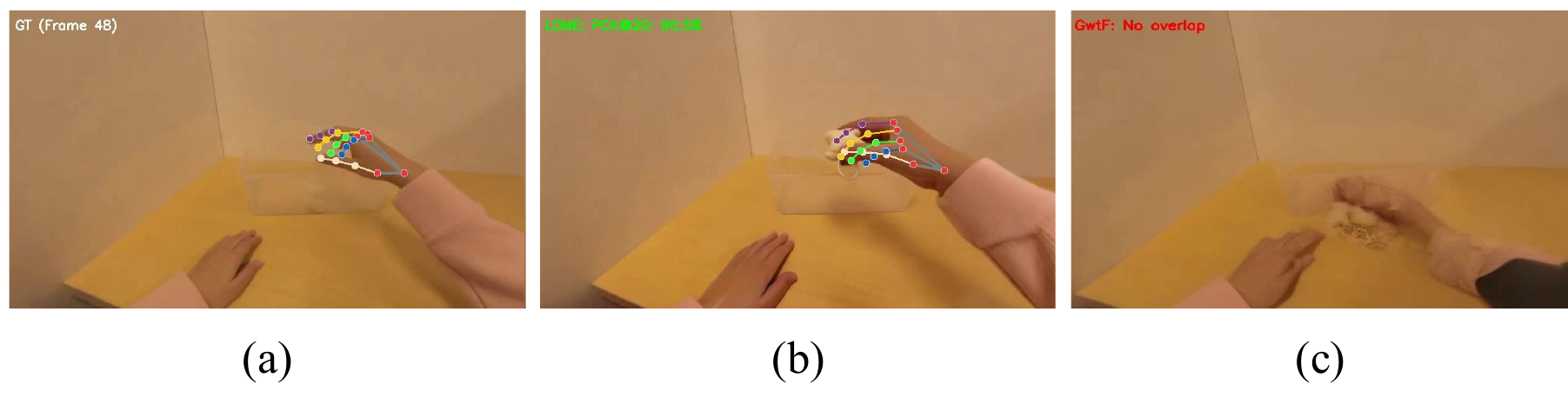} 
    \end{minipage}

    \vspace{-4mm}
    \caption{
     \textbf{Per-frame PCK@20 visualization.} We illustrate the PCK@20 evaluation by comparing detected hand locations in the generated videos against those in the ground-truth videos on a per-frame basis.}
    % \vspace{-4mm}
    \label{fig:pck_vis}
\end{figure}

\begin{figure}[ht]
    \centering 
    % First Subfigure
    % Second Subfigure
    \begin{subfigure}[b]{\linewidth}
        \centering
        \includegraphics[width=1.0\linewidth]{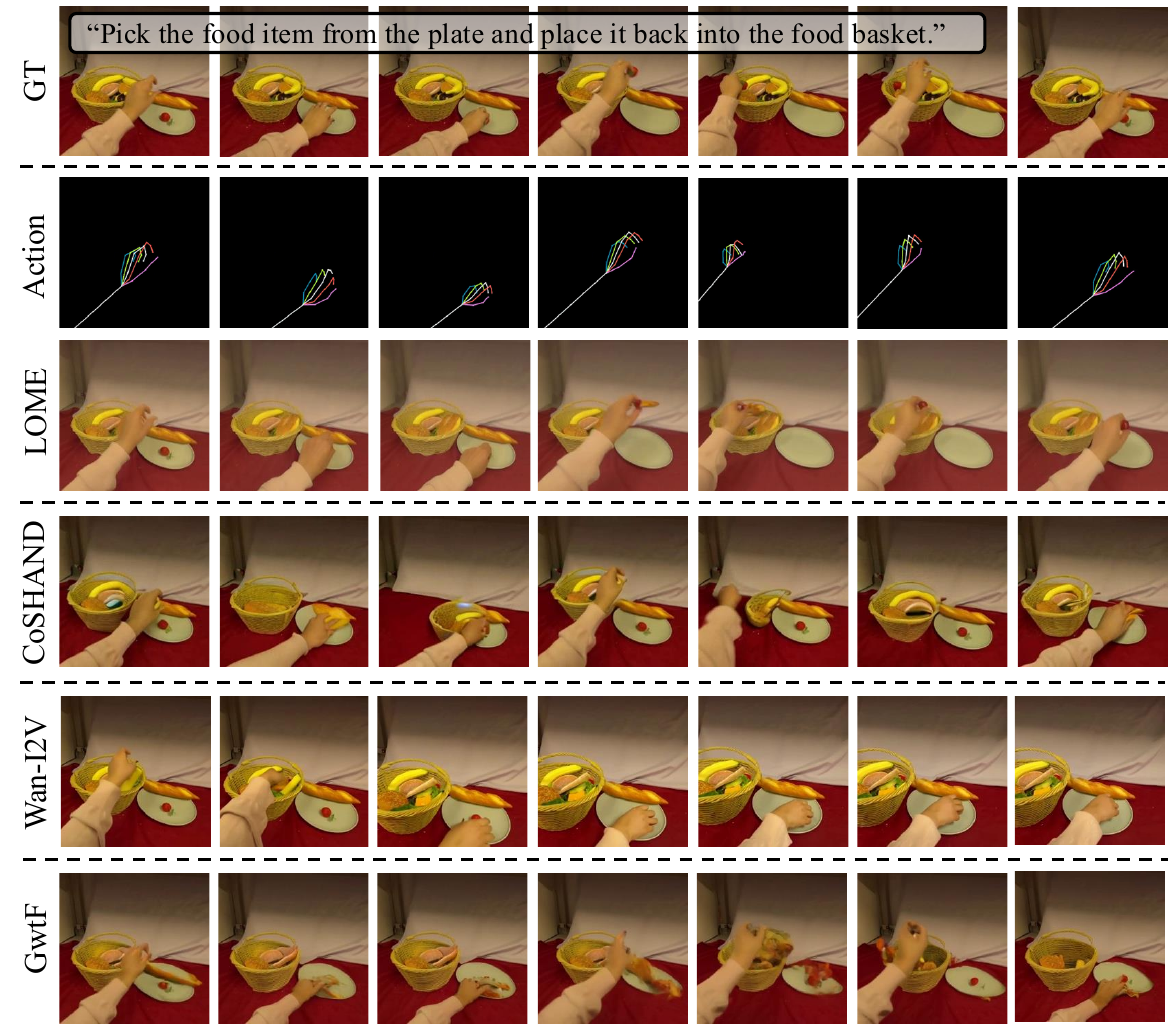}
        \vspace{-6mm}
        \caption{}
        \label{fig:qual3}
    \end{subfigure}
    
    \begin{subfigure}[b]{\linewidth}
        \centering
        % use \textwidth instead of \linewidth for the subfigure width to be safe
        \includegraphics[width=1.0\linewidth]{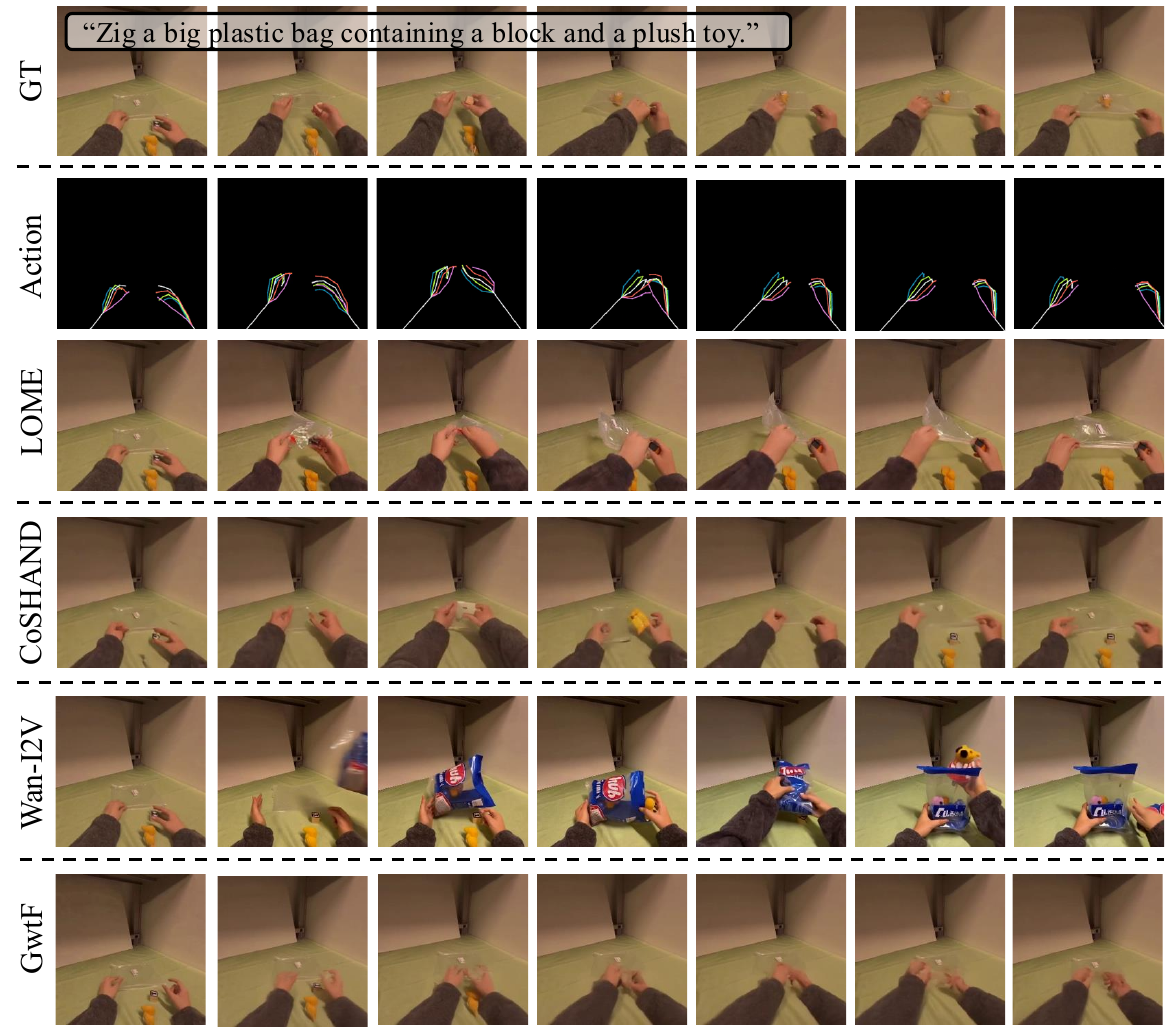}
        \vspace{-4mm}
        \caption{}
        \label{fig:qual1}
    \end{subfigure}
    \vspace{-6mm}
    % Main Caption for the whole figure
    \caption{\textbf{More qualitative comparisons across tasks.} We compare LOME (ours) with CoSHAND, Wan-I2V and GwtF on diverse human-object manipulations. “Action” denotes our 2D action maps; CoSHAND uses its own hand masks; Wan-I2V uses no action condition; GwtF uses GT optical flow as action condition. Text prompts are overlaid on GT video frames.}
    \label{fig:more_qualitative_comparison}
\end{figure}

\end{document}